\documentclass[10pt]{asme2ej}

\usepackage{epsfig} 
\usepackage{subfigure}
\usepackage{url}
\usepackage{amsmath}

\title{Performance Characterization of a Point-Cloud-Based Path Planner in Off-Road Terrain}

\author{Casey D. Majhor\thanks{Corresponding author.}
    \affiliation{
	Robust Autonomous Systems Laboratory\\
	Department of Electrical and Computer Engineering\\
	Michigan Technological University\\
	Houghton, Michigan 49931\\
    Email: cmajhor@mtu.edu
    }	
}

\author{Jeremy P. Bos
    \affiliation{
	Robust Autonomous Systems Laboratory\\
	Department of Electrical and Computer Engineering\\
	Michigan Technological University\\
	Houghton, Michigan 49931\\
    Email: jpbos@mtu.edu
    }
}

\begin{document}

\maketitle    

\begin{abstract}
{\it 
We present a comprehensive evaluation of a point-cloud-based navigation stack, MUONS, for autonomous off-road navigation. Performance is characterized by analyzing the results of 30,000 planning and navigation trials in simulation and validated through field testing. Our simulation campaign considers three kinematically challenging terrain maps and twenty combinations of seven path-planning parameters. In simulation, our MUONS-equipped AGV achieved a 0.98 success rate and experienced no failures in the field. By statistical and correlation analysis we determined that the Bi-RRT expansion radius used in the initial planning stages is most correlated with performance in terms of planning time and traversed path length. Finally, we observed that the proportional variation due to changes in the tuning parameters is remarkably well correlated to performance in field testing. This finding supports the use of Monte-Carlo simulation campaigns for performance assessment and parameter tuning.
}
\end{abstract}

\section{Introduction}
\begin{figure*}[ht!]
\centering
\subfigure[]{\label{fig:}\includegraphics[width=2.345in]{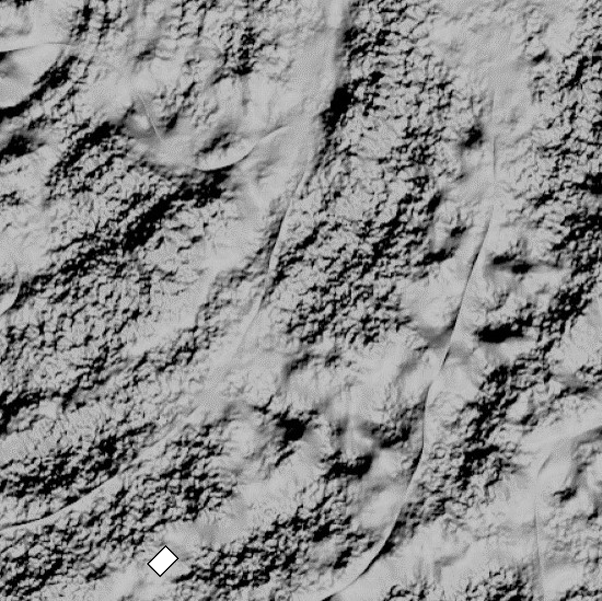}}
\subfigure[]{\label{fig:}\includegraphics[width=4in]{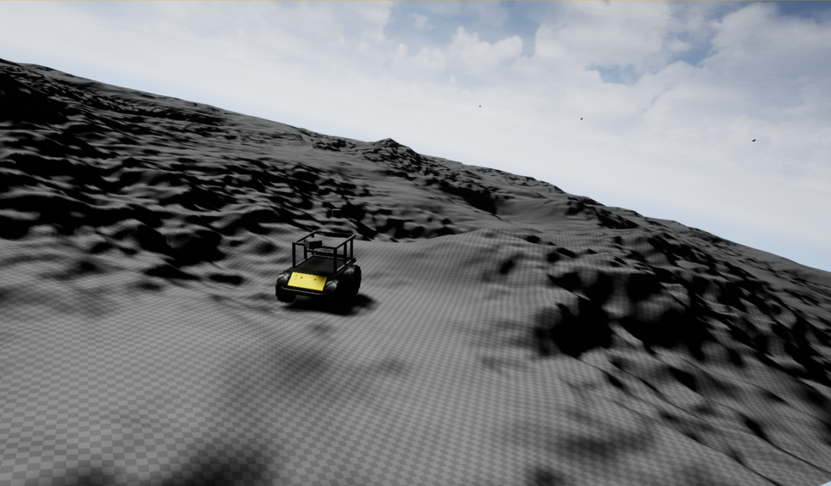}}
\caption{Top-down (a) and perspective views (b) of a multifractal terrain (\emph{terrain 1}) in Unreal Engine used in this work. The simulated autonomous ground vehicle in (b) is shown with its corresponding location shown in (a) as a white rectangle.}
\label{fig:T1_top_perspective}
\end{figure*}

Autonomous ground vehicle (AGV) navigation in off-road environments presents unique challenges not encountered in urban settings \cite{carruth2022challenges}. In contrast to urban areas, off-road terrains lack defined roads, complicating the identification of safe routes. Off-road terrains are often rough, uneven, and can contain impassable areas. Figure~\ref{fig:T1_top_perspective} illustrates a top and perspective view of a modeled AGV navigating such complex terrain. Failure to account for these complexities can result in AGVs becoming immobilized, either by getting stuck or rolling over \cite{darpa2004,thrun2006stanley}.
%

Most existing methods involve a path planner that searches the environment for a collision-free, low-cost path, typically across a 2D or 2.5D costmap \cite{overbye2021path}. This approach involves finding a sequence of cells connecting a starting pose to a goal. While effective in relatively flat terrains and indoor environments, costmaps cannot adequately represent terrain where the wrong approach angle violates the AGV’s kinematic constraints. These representations also have difficulties representing multi-level environments, pathways under bridges, or overhanging trees. Additionally, costmaps are often tailored to specific platforms and are not easily transferable between vehicles with different capabilities and constraints. Alternative map representations, such as meshes \cite{ruetz2019ovpc} or voxel grids \cite{ji2018adaptive}, require extra processing and algorithms to construct the planning map from sensor data. Point-cloud-based path planning addresses these limitations by leveraging lidar data to accurately represent the complexities of off-road terrain in three dimensions. Unlike meshes and voxel grids, this method does not require additional computation to reconstruct the terrain surface. While the viability of point-cloud-based path planning \cite{krusi2017driving} has been demonstrated, no attempt has been made to characterize performance across varied off-road terrains. 

Characterizing performance involves evaluating the ability to achieve a performance goal but also understanding performance variability. Performance metrics may include planning time, traversed path length, and success rate among others. Further, assessing performance across different navigation tasks, terrains, and algorithmic perturbations is essential to establish robustness. Exploring this varied-parameter space in field experiments is challenging due to the need for numerous trials to obtain reliable statistics. Field trials are also notoriously expensive, time-consuming, and prone to complications.

Characterizing the performance of path-planning algorithms can be complex. For instance, sample-based search algorithms exhibit non-deterministic performance in both planning time and path length, even in repeated navigation tasks. Furthermore, these algorithms have numerous tuning parameters. Varying the maximum search iterations, search radii, and step sizes between waypoints \cite{bhardwaj2020differentiable, binch2020context} and other parameters can all affect performance. These parameters, which may not always be accessible to an operator, are often manually tuned during field experiments to suit specific terrains or missions. The extent to which deviations from local or global optima impact performance is often unclear. 

Motivated by the "Driving on Point Clouds" (DoPC) work of Krüsi et al. \cite{krusi2017driving}, we aim to evaluate the reliability and performance of point-cloud-based planning and navigation in unstructured off-road terrains. Our approach to achieving this aim is a thorough characterization of our own point-cloud-based planning and navigation stack, MUONS (Michigan Technological University's Off-road Navigation Stack). MUONS is a terrain-aware approach that utilizes sample-based search methods for real-time path planning and re-planning, leveraging point-cloud data. While inspired by DoPC, MUONS employs different strategies for trajectory planning, map maintenance, and localization, representing a simplified variant focused on the path planning task. In this work, we assess the performance of MUONS across varying terrain types, mission scenarios, and tuning parameters through a combination of simulated and field experiments.

To characterize the performance of MUONS, we evaluate 30,000 simulated "A to B"  traversal tasks. The experiments are conducted over kinematically challenging off-road terrains generated using a multifractal approach \cite{majhor2024multifractal}. We select 20 combinations of MUONS's path planning tuning parameters for evaluation. The effect of parameter variations on performance metrics is assessed by comparing mission and parameter sets to the broader population statistics and by correlation. Our simulation results are validated by comparison to a limited set of field experiments evaluating a best, typical, and a worst-performing parameter set.

In simulation, the MOUNS-equipped AGV achieves a high success rate of over 0.98 over 30,000 trails. In the field, MUONS demonstrated a similar 1.0 success rate over 30 trials. The relative performance of the three parameter sets is also reflected in our field trials.  The large number of trials in our simulation experiment allows us to characterize performance statistically and suggest that both traversed path length (TPL) and planning time (PT) can be well approximated by exponential distributions. Simulation results also suggest that mission and parameter selection are more significant than terrain fractal roughness. Our correlation analysis suggests that the expansion radius of the bi-directional RRT initial planning step has the most significant impact on performance.  

The main contributions of this paper are as follows:
\begin{enumerate} \item Robustness of MUONS: We characterize our point-cloud-based off-road navigation stack (MUONS) over 30,000 A to B traversals and 30 field trials. Our simulation trials span 15 missions across 3 fractal-generated terrains and 20 sets of path planning parameters, allowing us to statistically characterize performance. MUONS achieved a success rate of 0.98 in simulation and had no failures in the field. 
\item Our simulation results indicate that both planning time and excess path length can be modeled by exponential distributions. 
\item The bi-directional RRT expansion radius is found to have the largest impact on performance among the tuning parameters considered. 
\item The variation in performance observed due to the choice of tuning parameters was found to agree exceptionally well between simulation and field testing. This supports the use of Monte-Carlo testing in simulation for both performance characterization and parameter tuning for off-road AGVs. It also suggests that extensive field testing campaigns can be replaced by more limited validation campaigns.  
\end{enumerate}

The remainder of this paper is organized as follows: Following this section, we review prior work related to off-road, point-cloud-based path planning. In Section~\ref{sec:prelims}, we present preliminary details of MUONS. Our experimental methods are outlined in Section~\ref{sec:methodology}. The results from simulation and field testing are discussed in Sections~\ref{sec:sim_results} and \ref{sec:field_results}, respectively. Finally, we conclude the paper and discuss future work in Section~\ref{sec:conclusions}.


\section{Related Work}

For autonomous navigation, the choice of map representation is crucial. Popular representations of off-road terrain for path planning include 2D and 2.5D costmaps~\cite{roy2018hierarchical}, meshes~\cite{ruetz2019ovpc}, and voxel grids~\cite{ji2018adaptive}. However, these approaches have notable limitations when navigating outdoor, rough terrains.

As a map representation, point clouds overcome these limitations by avoiding the need to explicitly reconstruct terrain from sensor data. Point-cloud planning is efficient because it utilizes already available point-cloud maps, which are often byproducts of simultaneous localization and mapping (SLAM) tasks~\cite{shan2018lego}. Unlike cost or traversability maps, point clouds are vehicle-agnostic, allowing them to be shared and used by different AGVs for path planning~\cite{chen2019cooper}.

Point clouds, as a map representation, offer significant advantages in off-road navigation. Building on these benefits, various approaches have been explored. For instance, point clouds generated from UAV images are utilized in an RRT-based global path planner for an AGV in~\cite{fedorenko2018global}. The traversability of the terrain is assessed by fitting points around a prospective waypoint to a plane to check if roughness, roll, and pitch constraints are satisfied. Planning experiments demonstrate that using UAV-derived point clouds is practical for this approach.

Point-cloud-based path planning methods for off-road navigation are a relatively recent development, with only a few existing approaches. One of the pioneering works in this domain is the three-step planner proposed by Krüsi et al.~\cite{krusi2017driving}, which serves as the foundation for our work. Their approach consists of initial path exploration using Bi-RRT, followed by optimization with rapidly-exploring random trees-star (RRT*), and a final local trajectory optimization (LTO) step. The effectiveness and practicality of their planner were demonstrated through offline planning evaluations and field tests in outdoor terrain, establishing it as a viable approach for off-road autonomous navigation.

Zhang et al.~\cite{zhang2019traversability}, also propose a method to plan on point clouds using a multi-step planner, similar to~\cite{krusi2017driving}. The first step is a non-holonomic A* algorithm which discovers an initial path while considering the vehicle kinematics, proximity to obstacles, and terrain traversability. This initial path consists of a sequence of 3D poses from start to goal. The path is then input to a trajectory optimization step which improves the path characteristics by considering a vehicle suspension model. 

In this work, we aim to address several gaps in the literature concerning point-cloud-based path planning. While existing research has focused on the viability of point-cloud-based planning, limited attention has been given to understanding the reliability and variability of performance. This is particularly crucial when employing sample-based planning methods, where performance can vary due to their stochastic nature. Furthermore, planning algorithms often involve numerous tuning parameters that can significantly impact performance~\cite{bhardwaj2020differentiable, binch2020context}. Few studies have explored the sensitivity of performance to these parameters and whether they can and should be optimized for different navigation tasks and terrains. Understanding how these parameters affect performance is essential for efficient tuning. Another gap is the extent to which planner performance generalizes across different versions of the same mission and between terrain types. 
\section{Preliminaries} 
\label{sec:prelims}
In this section, we outline how MUONS plans and tracks global paths given start and goal locations. First, we provide a high-level overview of our planning and navigation stack. Next, we describe the path-planning parameters explored in this work. Finally, we present an overview of our simulation environment and explain how localization and mapping are performed both in simulation and in the field.

\subsection{MUONS Overview}
\label{sec:muons_arch}
MUONS, the Robot Operating System (ROS) based 3D path planner used in this work, is based on work by Kr{\"u}si et al.~\cite{krusi2017driving}. Our original implementation of MUONS was developed around a Clearpath A200 Husky AGV, but it has also been extended to platforms with complex suspensions \cite{jeffries2021muons} in simulation.

MUONS employs a three-phase path planning process. The first phase utilizes a start/goal Bi-RRT algorithm to quickly find a kinematically feasible path between the input start and goal poses. The path found via Bi-RRT is then passed to a RRT* algorithm to optimize the path length. Finally, the RRT*-optimized path is input to a LTO, which further refines the path for traversability, curvature, and length. Each path-planning step leverages terrain assessment (TA) to ensure the AGV's pose satisfies roll and pitch constraints. The TA infers the AGV's pose at each potential waypoint by estimating a terrain surface normal using a robot-sized patch of points. Additionally, our TA measures and checks the roughness of the terrain against a user-defined limit to ensure the path is suitable for the AGV.

MUONS uses a pure-pursuit path tracker, adapted from Snider's method~\cite{snider2009automatic}, to control the AGV's heading via angular velocity. In this approach, the AGV tracks a look-ahead point on the desired path. This look-ahead point is a fixed distance from the AGV and is updated incrementally as the AGV moves toward it.

\subsection{Path-Planning Parameters}
\label{sec:path_planning_params}
Understanding the impact of path-planning parameters on performance metrics, such as planning times and traversed path lengths, is crucial for efficient tuning. By adjusting these parameters, we can gain insights into performance sensitivities and robustness. Since we assess performance across various "A to B" navigation tasks, each requiring numerous trials, we limit our experiments to 20 combinations of seven parameters, summarized in Table~\ref{tab:params}. These specific parameter-value combinations are referred to as \emph{parameter sets} (PS) ~\cite{kysar2021unstructured}.

\begin{table}[!b]
\caption{MUONS path planning parameters with descriptions, selected for experimentation}
\label{tab:params}
\centering
\resizebox{\columnwidth}{!}{%
\begin{tabular}{l l}
\hline
Parameter & Description \\
\hline
$r^{bi-rrt}_{exp}$ & Bi-RRT expansion radius \\
$k$ & Start to Goal RRT* radius proportional constant  \\
$r^{rrt*}_{exp}$ & RRT* distance offset from near vertex  \\
$\alpha$ & Weighting of vertex count for RRT* exit  \\
$N^{avg}_{max}$ & Maximum vertices in radius for RRT* exit \\
$\delta_{max}$ & Maximum local trajectory optimization offset  \\
$\delta_{min}$ & Minimum local trajectory optimization offset \\
\hline
\end{tabular}%
}
\end{table}

The seven parameters are associated with the initial Bi-RRT path planning, RRT* refinement, and LTO. In this first step, $r^{bi-rrt}_{exp}$, defines the maximum distance each search tree expands towards randomly sampled waypoints in the point-cloud map. In the RRT* phase of path planning, $k$, defines the search radius, $r_{near}$, around a selected RRT tree vertex for sampling a new waypoint. The $r_{near}$ radius is defined as
\begin{equation}
\begin{aligned}
r_{near} = k \, dist(v_{start}, v_{goal})
\label{eqn:r_radius}
\end{aligned}
\end{equation}
where $k$ is a constant of proportionality and \emph{dist($v_{start}$, $v_{goal}$)} represents the Euclidean distance between the start and goal vertices, denoted as $v_{start}$ and $v_{goal}$, respectively.
 
The third parameter, also in the RRT* step, $r^{rrt*}_{exp}$,  defines the maximum expansion radius towards a sampled point for the RRT* phase of planning. Besides reaching a maximum number of iterations, RRT* can terminate by sufficiently exploring the area around the initial Bi-RRT path. This exploration criterion is defined by calculating the moving average ($N^{avg}_i$) number of tree vertices (vertex density) within a radius of a newly sampled point in each iteration $i$. The moving average is defined as
\begin{equation}
\begin{aligned}
N^{avg}_i = \alpha N_i + (1 - \alpha)N^{avg}_{i-1}, \quad 0 < \alpha \ll 1 
\label{eqn:n_i_avg}
\end{aligned}
\end{equation}
where $\alpha$ is the fourth parameter considered. In equation~\ref{eqn:n_i_avg}~\cite{krusi2017driving}, $\alpha$ is a weighting factor, $N_i$ is the current sample density, and $N^{avg}_{i-1}$ is the previous iteration's average vertex density. If $N^{avg}_i$ exceeds a chosen threshold parameter, $N^{avg}_{max}$, RRT* terminates; $N^{avg}_{max}$ is our fifth parameter.

In the LTO phase of path planning, $\delta_{max}$ defines the initial lateral offset distance on the left and right sides of each vertex, from which a new graph structure is created for optimization. At each LTO iteration, a randomly selected vertex is chosen, and the offset of its subvertices is reduced. The LTO terminates when an offset of $\delta_{min}$ is reached at each vertex or when a maximum number of LTO iterations is reached. We choose $\delta_{max}$ and $\delta_{min}$ as our sixth and seventh parameters.

\subsection{Simulation Environment}
To be useful for predicting performance in the field vehicle-terrain interactions in simulation need to be modeled with sufficient accuracy. In a prior work~\cite{young2020unreal} we developed a simulation environment in the Unreal Engine (UE)~\cite{unrealengine_misc}, capable of modeling these interactions. This UE architecture includes a simulated Clearpath Husky AGV~\cite{husky} and uses NVIDIA's PhysX~\cite{physx} physics engine subsystem. PhysX has been shown to accurately model vehicle-terrain interactions~\cite{erez2015simulation} crucial to this work. Also modeled in simulation are multiple sensor sources we use for mapping and localization. Sensors include lidar, inertial measurement units (IMUs), wheel odometry, and GNSS. 
\subsection{Localization and Mapping}
\label{sec:local_and_map}
Vehicle localization in simulation is achieved using the ROS \emph{robot\_localization} package~\cite{MooreStouchKeneralizedEkf2014,gibb2018nondestructive, macenski2020marathon}. In this approach, two extended Kalman filters (EKFs) are employed. The first EKF takes simulated continuous sensor data, such as wheel odometry and IMU measurements, as input to obtain a locally accurate state estimate. This state estimate is then fused with a second EKF instance that incorporates both simulated continuous and discrete data, such as GPS measurements, to achieve an accurate global state estimate.

For field testing MUONS, we localize using real-time kinematic GNSS/INS data. This data is input to three EKF instances, corresponding to the world, local, and odometry reference frames. As in the simulation, the ROS \emph{robot\_localization} package is used to implement the EKFs. This approach provides high-frequency position estimates with centimeter-level accuracy. Our intent in both simulation and field testing is to remove localization as a variable to the extent possible. 

Point-cloud maps in simulation are generated by pushbrooming a virtual lidar sensor acorss the terrain and then filtering to produce a map with consistent sampling. In the field, we use LiDAR Inertial Odometry via Smoothing and Mapping (LIO-SAM)\cite{shan2020lio} for mapping. LIO-SAM is a LiDAR-based simultaneous localization and mapping approach that fuses tightly-coupled IMU data with LOAM-based lidar features\cite{zhang2014loam,cattaneo2022lcdnet, ebadi2023present}. With the Husky's sensing and localization algorithms active, the AGV is teleoperated around the operational area while recording data in a ROS bag. This recorded data is then input into LIO-SAM via bag playback to create a point-cloud map.

\section{Methodolgy} 
\label{sec:methodology}
We select three performance measures for use in this work:
\begin{itemize}
	\item \textbf{Success Rate (SR):} The rate at which the AGV successfully reaches the goal location and serves as a measure of reliability. In some cases, it is more instructive to use the failure rate (FR) defined as $FR = 1 - SR$.
 	\item \textbf{Traversed Path Length (TPL):} The path length traversed by the AGV while following the planned path. Results are presented in terms of the excess path length, $\Delta x$, compared to the Euclidean distance between a start and goal.
	\item \textbf{Planning Time (PT):} The time taken by MUONS to plan a path in integer seconds.
\end{itemize}

To validate the simulation results, we selected a best, typical, and worst-performing parameter set from the simulation experiments for validation in the field. These parameter sets were chosen by evaluating each parameter set using a cost function incorporating our performance metrics.
\subsection{Path-Planning Parameter Selection}
\label{sec:param_selection}
As described in Section~\ref{sec:path_planning_params}, we focus on a subset of seven path-planning parameters involved in MUONS. We consider 20 combinations of these seven parameters, sampled randomly from ranges and distributions consistent with the values published in DoPC~\cite{krusi2017driving}. Table~\ref{tab:param_dist} summarizes the ranges and distributions, while Table~\ref{tab:param_sets} in the appendix explicitly lists the 20 parameter set values.

\begin{table} [b!]
\caption{MUONS path planner parameter ranges and distributions used for random selection}
\begin{center}
\label{tab:param_dist}
\begin{tabular}{l c l}
\hline  
Parameter & Distribution Type  & Distribution Notes \\ \hline
$r^{bi-rrt}_{exp}$   & [0.03-3.0]          & Uniform            \\  
$k$         & $\mu$=$\sigma$=0.1  & Gaussian and positive          \\  
$r^{rrt*}_{exp}$                  & $\mu$=$\sigma$=0.1  & Gaussian and positive         \\  
$\alpha$                   & $\mu$=$\sigma$=0.25 & Gaussian and positive           \\  
$N^{avg}_{max}$               & [5-100]             & Uniform            \\  
$\delta_{max}$              & [0.01-1.0]          & Joint uniform with   \\  
                        &                       & $\delta_{min}$       \\
$\delta_{min}$              & [0.001-0.1]         & Joint uniform with and   \\  
                        &                       & greater than $\delta_{max}$ \\
\hline
\end{tabular}
\end{center}
\end{table}

We consider that the AGV used in this work is slightly smaller than the one used in DoPC. For this reason, we sample \( r^{bi-rrt}_{exp} \) with a uniform distribution between 0.03 and 3~m. This range is chosen to include the value of 0.6~m used in DoPC, allowing for selection both above and below it.

The parameter $k$ is sampled from a distribution with a mean of $0.1$~m, resulting in an $r_{near}$ distribution mean of $3.5$~m, since our traversal experiments in simulation have a Euclidean distance of $35~m$. This is close to the equivalent value of $3.6~m$ used in \cite{krusi2017driving}. The RRT* expansion radius, $r^{rrt*}_{exp}$, is sampled from the same distribution, with the same mean and standard deviation as $k$. This sampling approach is more conservative compared to the expansion radius of $1.8$~m selected in \cite{krusi2017driving}.

Although $\alpha$ is not specified in \cite{krusi2017driving}, we sample it from a Gaussian distribution with a mean and standard deviation equal to $0.25$. The parameter $N^{avg}_{max}$ is sampled from a uniform distribution between $5$ and $100$, which includes the value of $30$ used in \cite{krusi2017driving}. The ranges and distributions for $\delta_{max}$ and $\delta_{min}$ are chosen based on the respective values of $0.08~m$ and $0.04~m$ used in \cite{krusi2017driving}.

\begin{figure*}[t!]
\centering
\subfigure[]{\label{fig:M1_5rand}\includegraphics[width=2.19in]{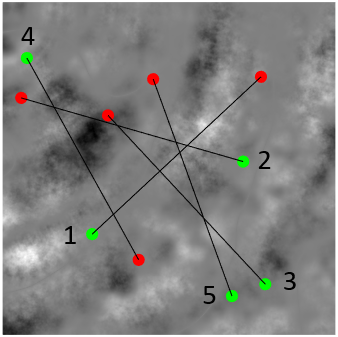}}
\subfigure[]{\label{fig:M2_5rand}\includegraphics[width=2.19in]{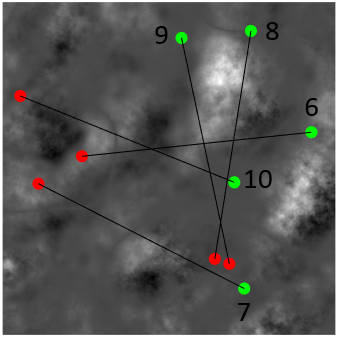}}
\subfigure[]{\label{fig:M3_5rand}\includegraphics[width=2.19in]{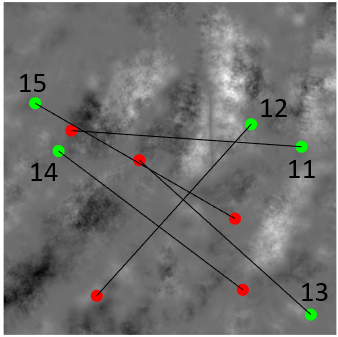}}
\caption{$15$ randomly selected navigation tasks with start/goal locations shown in the digital elevation maps used to generate terrains 1 (a), 2 (b), and 3 (c). The start locations are labeled by their task number and connected to their respective goal with a line. Note that darker and brighter shades represent lower and higher elevations, respectively.}
\label{fig:T1-3_rand_missions}
\end{figure*}

\subsubsection{Fixed Parameters}
Throughout all phases of path planning, the geometry of the terrain is assessed to quantify its traversability. The pose of the AGV at a prospective waypoint is estimated by fitting a plane to a robot-sized volume of points at each waypoint. If the estimated pose of the AGV exceeds roll limits of $(-0.523, 0.523)$ radians or pitch limits of $(-0.61, 0.785)$ radians, the waypoint is deemed untraversable.

Each point in the robot-sized patch of points is assigned a roughness score. At each point, a normal plane is fitted using the points within a given radius. The roughness score for each point within the radius of a selected waypoint is calculated as the largest step height between any two points used for fitting the normal plane. This step height is measured along the direction of the normal plane. The waypoint is considered untraversable if any point-roughness score exceeds a user-defined limit of 0.19~m, equal to the AGV tire radius considered in this work.

\subsection{Terrain Generation}
We use a multifractal technique~\cite{majhor2024multifractal} to generate the off-road terrain maps used in simulation. This approach allows us to create randomized terrain maps with similar roughness statistics. As demonstrated in~\cite{majhor2024multifractal}, the ability to generate different maps with the same statistics ensures that our results generalize across similar terrain types rather than being associated with a specific map. It also allows us to tune the roughness statistics so that traversing the terrain is challenging but not so difficult that a feasible path is unlikely to exist.

We use this method to create three unique terrains for use in this work, referred to as \emph{terrains 1, 2,} and \emph{3} (T1, T2, and T3). T1 and T2 are generated with a Weierstrass-Mandelbrot (W-M) fractal coefficient of $2.45$ but with different random seeds. This results in terrains with similar roughness but different features. The fractal coefficient for Terrain 3 is 2.6 resulting in a rougher terrain. In our work introducing this technique\cite{majhor2024multifractal} we demonstrated that increasing this coefficient from $2.45$ to $2.6$ reduced the SR for straight-line paths from 0.75 to 0.5.

\subsection{Random Navigation Task Selection}
\label{sec:rand_mission_selection}
In our experiments, MUONS is applied to 15 start-to-goal navigation tasks or missions. The start and goal locations are randomly selected, ensuring a Euclidean distance of $35$ m. To ensure the feasibility of the start and goal locations, we restrict them to categorized low- or semi-rough terrain. We categorize the terrain by calculating the local elevation gradient of each DEM cell and filtering the resulting gradient map using empirically determined thresholds to identify low-, semi-, and high-roughness areas. Starting locations are constrained to low-roughness terrain to prevent vehicle movement before traversal, while goal locations can be in either low or semi-rough terrain to ensure reachability. Figure\ref{fig:T1-3_rand_missions} illustrates the 15 sets of start and goal locations defining the navigation tasks, with green and red dots representing the start and goal locations, respectively, connected by a line.

\subsection{Evaluation Metrics}
\label{sec:eval_metrics}
Our 30,000 simulation experiments are conducted over 15 missions on three terrain maps. For each mission (M1 to M15), each PS (PS1 to PS20) is evaluated 100 times. We refer to the population of all experiments as PA. Performance in each mission and each parameter set are compared statistically to this population. Missions 1 to 5 (M1 to M5) are considered P1, M6 to M10 are conducted on T2 and comprise P2. Missions 11 to 15 on T3 belong to P3. Because the terrain statistics across T1 and T2 are the same, the statistics of P1 and P2 can be analyzed together as P12. In analyzing our results, we observed a single mission, M2, to be an outlier so we also analyze the population statistics excluding this mission as PA$^{\prime}$, P12$^{\prime}$, and P1$^{\prime}$ for the populations representing all missions and parameters sets, those for T1 and T2, and T1 only respectively. When evaluating planning time we also found it useful to exclude PS18 from PA$^{\prime}$ since it is an outlier, and we refer to this population as PA$^{\prime\prime}$. 

Performance in terms of SR, TPL, and PT are evaluated across each of these populations, and the effect of terrain, mission, and tuning parameters is considered. We also examine the relative benefits of tuning for a specific mission or terrain by comparing best performing PS to those performing the best across the larger population (PA, P12, etc). In assessing the correlation between performance metrics and tuning parameters we employ the Spearman partial correlation coefficient. In some cases, the failure rate (1- SR) is used in place of SR to better visualize variations in performance.

Repeating our simulated test plan in the field for a perfect comparison would be prohibitive. Instead, we select from 3 parameter sets we score as "best", "worst", and "typical" via a cost function incorporating three performance metrics as
\begin{equation}
    Rank = 1-FR-\hat{\Delta x}_{med}-\hat{\Delta x}_{IQR}
    \label{eq:rank}
\end{equation}
where the $\hat{\Delta x}_{med}$ and $\hat{\Delta x}_{IQR}$ are the median and IQR of the excess path length, $\Delta x$, normalized by the start-to-goal Euclidean distance. Our aim in our field trials is to verify that the performance is similarly correlated to simulation. We found that performance in PT is mostly independent of mission and terrain type and is, therefore, not included in our ranking.  
\section{Simulation Experiments}
\label{sec:sim_results}
This section presents the results of our simulation experiments. Evaluation is performed over three synthetic terrains. For each terrain, 5 traversal tasks of uniform length are selected randomly. MUONS is tasked with navigating the AGV from the start to the goal position 100 times for each of the 20 parameter sets. As a result, 10,000 traversals are performed on each terrain map, totaling 30,000 traversals presented in this section. Each PS is evaluated 1,500 times and each mission 2,000 times.
\subsection{Simulation: Success Rate}
\begin{figure}[h!]
\begin{centering}
\subfigure[]{\includegraphics[width=3 in]{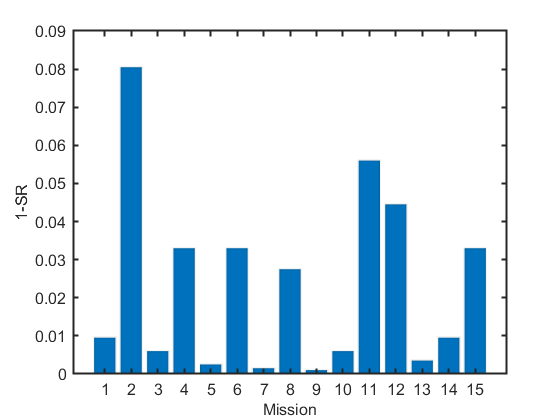}}
\subfigure[]{\includegraphics[width=3 in]{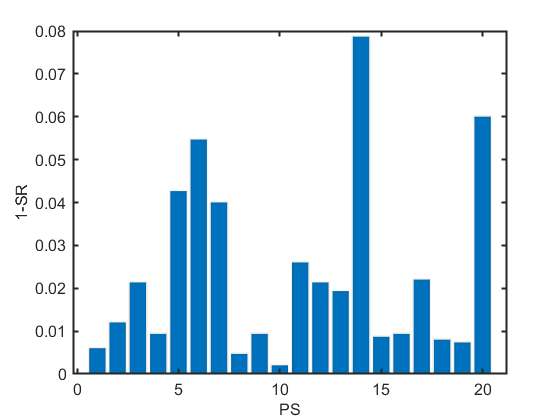}}
\caption{MUONS success visualized using the FR (1-SR) by mission (a) and PS (b). Each mission represents 2,000 trials and each PS 1,500 trials.}
\label{fig:SR_by_M_PS}    
\end{centering}
\end{figure}

In our previous work \cite{majhor2024multifractal} we demonstrated that the mission SR across straight-line paths over the fractal-generated terrain maps used here can be as low as 0.4. For terrain maps with statistics similar to T1 and T2 the straight-line SR was $0.75$ and for terrains similar to T3, the SR was 0.5; the aggregate success rate was $0.65$. When using MUONS for path planning the aggregate success rate (PA) increases to $0.98$ representing $694$ failures over the 30,000 trials. Of these, $401$ occurrences were in terrain maps T1 and T2 (P12) and $293$ in T3 (P3) representing success rates of $0.98$ and $0.97$ for P12 and P3 respectively. These results represent an improvement of $0.23$ (P12) and $0.47$ (P3) compared to the straight-line path.  Results over missions and PS are detailed in Figure \ref{fig:SR_by_M_PS} in terms of the $FR = 1 - SR$ to aid visualization. 

Examining Figure \ref{fig:SR_by_M_PS} (a) we observe that the M2 has the highest failure rate at $0.08$ (SR $0.92$) representing a total of $161$ failures or 26\% of the total over PA and 48\% in P12. For this reason, we often exclude M2 from our analysis as an outlier. When excluded the SR for PA$^{\prime}$ increases slightly (0.981 from 0.977), the SR for P12$^{\prime}$ increases to 0.99. By contrast, the success rate for M9, the highest in PA, was 0.999 representing only 2 failures across 2000 trials. 

To estimate the effect of terrain on SR we first compare T1 and T2. Here, we would expect performance to be similar given the two maps were generated with the same fractal statistics. Instead, a total of 263 failures occurred for T1 and 138 for T2. However, as pointed out above 161 of these failures are attributable to M2. Excluding this mission and comparing SR yields 0.987 for P1$^{\prime}$ and 0.986 for P2 and 0.987 for P12$^{\prime}$. Comparing this value to the SR for T3 we observe a decrease between 0.01 and 0.02 for the rougher terrain. We interpret this finding as indicating MUONS is effective at finding and navigating kinematically challenging terrain assuming a safe path exists regardless of terrain roughness. 

Figure \ref{fig:SR_by_M_PS} (b) shows the failure rate by PS over 1,500 trials, 100 each for 15 missions. PS14 has the most failures with 102 (SR 0.932) and PS10 the least with only 3 (SR 0.998). The span between the best and worst values is 0.066 suggesting that the choice of parameter is more important than terrain, again assuming a safe path exists. Comparing these parameters to the overall population success rate further suggests tuning can reduce the SR by as much as 0.04 or increase performance by 0.02 compared to a parameter set chosen at random. 

\subsection{Simulation: Traversed Path Length}
\label{sec:TPL_in_sim}
\begin{figure} [b!]
\centerline{\includegraphics[width=3in]{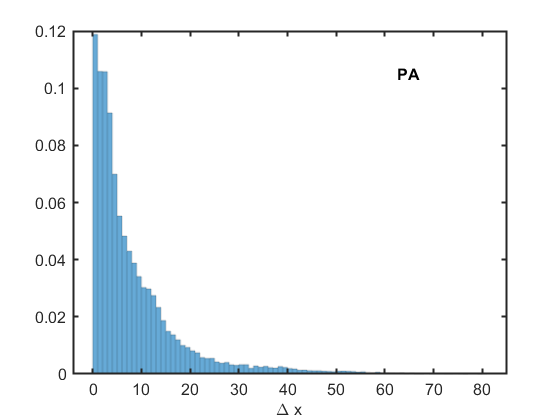}}
\caption{Normalized histogram of excess path length, $\Delta x$, in meters relative to the Euclidean distance of 35 m for the entire simulation population (PA) of 30,000 trials. The distribution is well-characterized by an exponential distribution with a rate parameter of $1/\lambda_{TPL,PA} =$ 8.9 m.}\label{fig:TPL_PA}
\end{figure}
As we mentioned in Section \ref{sec:methodology} the straight-line distance between each set of start and goal locations was a simulated 35 m. In the previous section, we highlighted that applying MUONS increases the SR for a given mission between 23 and 47\% depending on the terrain roughness. A key finding of our previous work \cite{majhor2024multifractal} was that fractal roughness decreases SR along with negatively impacting other measures of performance. However, there is necessarily a cost associated with applying MUONS both in terms of the time required to plan the path and the overall path length or traversal time. The latter performance metric is the subject of this section. 

We choose TPL as a proxy for both the time to complete a mission and the effectiveness of the planner in finding a suitable path. The two metrics are effectively equivalent in this case because the Husky A200 platform, both in simulation and field testing, operates at a near-constant velocity across traversals. However, in contrast to the planned path length, TPL includes the effect of the pure pursuit controller. We feel this is an important consideration as we intend MUONS to plan paths that can easily be followed by the robot.

As before we start by examining performance metrics for the entire simulation population and then drill down into considering the effects of terrain, mission, and PS. A histogram of the excess traversed path length, $\Delta x$, is shown in Figure \ref{fig:TPL_PA} as an empirical PDF, where counts are normalized to better allow comparison between populations. Here, we observe that PA is well approximated by an exponential distribution with a rate parameter $1/\lambda_{TPL,PA} = 8.6$ m. This is not the case for smaller subsections of the population. So, it will be more useful to compare the population median and inter-quartile range (IQR), particularly for long-tailed distributions. Here for the full simulation population (PA) the median excess TPL is 5.1 m and the IQR 9.0 m. As for the SR, the population is skewed by M2, if that mission is excluded $1/\lambda_{TPL,PA^{\prime}}  = 6.9$ m. The median and IQR over PA$^{\prime}$ are 4.7 and 7.7 m as indicated in Table \ref{tab:summary}.

\begin{table*}[h!]
\begin{center}
\caption{Summary of population statistics from simulation experiments}
\label{tab:summary}
\begin{tabular}{l| c c c c c c c c} 
\hline
\hline
Metric	&	PA	&	T12	&	T3	&	T1	&	T2	&	T12$^{\prime}$	&	T1$^{\prime}$		&	PA$^{\prime}$	\\	\hline
SR	&	0.977	&	0.98	&	0.971	&	0.974	&	0.986	&	0.987	&	0.987	&	0.981	\\	\hline
TPL Mean (m)&	8.6	&	13	&	6.4	&	13	&	6.4	&	7.1	&	8.1	&	6.9	\\	\hline
TPL Media (m)	&	5.1	&	8.8	&	13.5	&	8.8	&	3.9	&	4.9	&	6.4	&	4.6	\\	\hline
TPL IQR	(m)&	9	&	13.5	&	6.2	&	13.5	&	7.3	&	8.46	&	8.9	&	7.7	\\	\hline
PT Mean	(s)&	53.65	&	59.12	&	42.69	&	76.48	&	41.75	&	54.02	&	69.35	&	35.15	\\	\hline
PT Median (s) &	30	&	32	&	24	&	38	&	27	&	30	&	35	&	27	\\	\hline
PT IQR (s)&	35	&	40	&	27	&	48	&	29.5	&	35	&	41	&	28	\\	\hline
\hline
\end{tabular}
\end{center}
\end{table*}

Examining Table \ref{tab:summary} and comparing populations between terrain types we observe that even excluding M2, between P1$^{\prime}$ and P2, the means differ by 2.7 m, on the other hand, P2 and P3 have nearly the same mean, median and IQR. We interpret this result as implying that T1 includes features that are more difficult to avoid or require careful planning. The similarity between P2 and P3 also hints that under some conditions MUONS performance in terms of TPL may be invariant relative to terrain roughness.

\begin{figure}[b!]
\centerline{\includegraphics[width=3in]{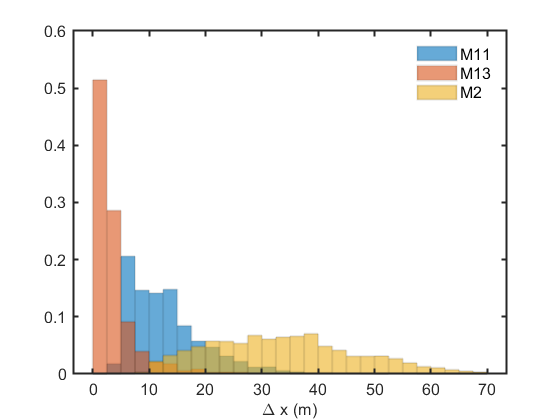}}
\caption{Normalized histogram of TPL in excess of 35 m, $\Delta x$, for three missions. M13 has the smallest median $\Delta x$ and M2 the largest. Because M2 is considered an outlier M11 (the next worst) is also included.}
\label{fig:TPL_M}
\end{figure}

In Table \ref{tab:tpl_m} the excess TPL, $\Delta x$ is presented for each mission. Once again, M2 stands out with a median $\Delta x$ of 33.9 m compared to 4.6 m for the simulation population (PA$^{\prime}$) excluding M2. The IQR is also 18.6 m compared to 7.7 m for PA$^{\prime}$. This IQR is nearly 10 m larger than the next largest (M11 at 8.8 m) underscoring the extent to which this mission is an outlier. Examining Figure \ref{fig:TPL_M} we also observe that the distribution of $\Delta x$ for M2 is two-sided around the mean. For all other missions, the distribution of excess path length is one-sided. MUONS performs the best in planning M13 where both the median and IQR are around half that of the PA$^{\prime}$. Also in Figure \ref{fig:TPL_M}, the distribution of TPL for M13 is roughly exponential while M11 can be better described as somewhat log normal. Considering M11 and M13 as the worst and best case in this context, the choice of mission can increase or decrease the median $\Delta x$ by 50\%. Also, for missions where MUONS performs well the IQR decreases 60\% compared to all possible missions. For difficult missions TPL variability increases by 14\% compared to PA$^{\prime}$.

\begin{table}[b!]
\begin{center}
\caption{TPL summary statistics by mission in terms of excess path length, $\Delta x$, in meters compared to the straight-line distance (35 m). The largest values across missions are in bold and the smallest are underlined.}
\label{tab:tpl_m}
\begin{tabular}{l| c c c} 
\hline
\hline
Mission	&	Mean	&	Median	&	IQR	\\	\hline
1	&	7.9	&	6.1	&	6.2	\\	\hline
2	&	\textbf{33.9}	&	\textbf{33.2}	&	\textbf{18.6}	\\	\hline
3	&	5.5	&	3.1	&	6.9	\\	\hline
4	&	14.0	&	12.6	&	6.7	\\	\hline
5	&	5.1	&	3.4	&	5.7	\\	\hline
6	&	5.0	&	\underline{2.3}	&	6.9	\\	\hline
7	&	6.1	&	4.9	&	6.9	\\	\hline
8	&	8.9	&	7.3	&	8.1	\\	\hline
9	&	4.1	&	2.0	&	4.8	\\	\hline
10	&	7.9	&	4.2	&	7.2	\\	\hline
11	&	13.2	&	11.9	&	8.8	\\	\hline
12	&	6.3	&	5.2	&	5.3	\\	\hline
13	&	\underline{3.5}	&	2.4	&	\underline{3.0}	\\	\hline
14	&	3.8	&	3.0	&	2.8	\\	\hline
15	&	5.2	&	3.5	&	4.1	\\	\hline
\hline
\end{tabular}
\end{center}
\end{table}

\begin{figure} [b!]
\centerline{\includegraphics[width=3in]{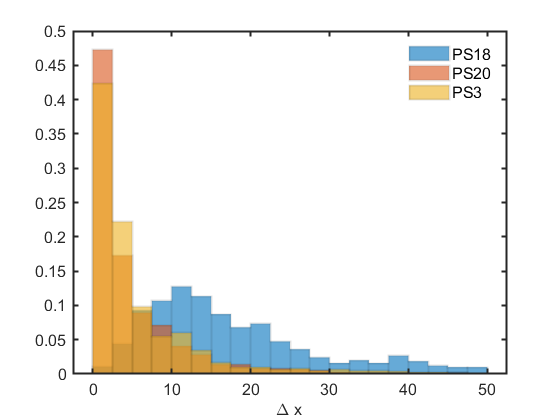}}
\caption{Normalized histogram of $\Delta x$ for three parameter sets. PS18 has the worst overall performance while P20 has the best. Because PS20 has a low SR (0.94) PS3 is included as the next best.}
\label{fig:TPL_PS}
\end{figure}

Next, we aim to understand the effects of parameter tuning in the context of these other variables. Table \ref{tab:tpl_ps} describes the mean, median, and IQR for the 20 parameter sets across all missions (PA). Here we see that among the PS there is also a clear outlier in PS18. The median value of $\Delta x$ is more than three times the median across PA$^{\prime}$ and the IQR is nearly double the baseline. PS20 performs the best among parameter sets reducing the median $\Delta x$ by 46\% and the IQR by 22\%. Unfortunately, the SR of PS20 is only 0.94. For this reason, we opt to use PS3 as the overall best and include it along with PS18 and PS20 in Figure \ref{fig:TPL_PS}. The median and IQR of $\Delta x$ associated with PS3 are 3.0 and 5.9 m, a reduction of 35\% and 23\% respectively compared to the baseline. While on a percentage basis these improvements seem substantial the actual improvement in these quantities is on the order of 1 to 2 m or 3 to 5\% of the straight-line path length. On the other hand, a poor choice of PS may increase the planned path length 10 m or more. 
%
\begin{table}[b!]
\begin{center}
\caption{TPL summary statistics by PS in terms of excess path length, $\Delta x$, in meters compared to the straight-line distance (35 m). The largest values across PS are in bold and the smallest are underlined.}
\label{tab:tpl_ps}
\begin{tabular}{l| c c c} 
\hline
\hline
PS	&	Mean	&	Median	&	IQR	\\	\hline
1	&	12.6	&	8.7	&	11.6	\\	\hline
2	&	9.5	&	6.1	&	9.3	\\	\hline
3	&	5.9	&	3.0	&	5.9	\\	\hline
4	&	11.5	&	8.1	&	10.2	\\	\hline
5	&	\underline{5.0}	&	2.6	&	\underline{5.6}	\\	\hline
6	&	5.6	&	2.9	&	6.1	\\	\hline
7	&	7.7	&	4.8	&	8.3	\\	\hline
8	&	8.0	&	4.6	&	7.8	\\	\hline
9	&	7.0	&	3.8	&	6.4	\\	\hline
10	&	11.2	&	8.1	&	9.2	\\	\hline
11	&	6.1	&	3.3	&	6.5	\\	\hline
12	&	7.0	&	3.5	&	7.1	\\	\hline
13	&	8.0	&	4.7	&	8.6	\\	\hline
14	&	8.9	&	6.2	&	8.7	\\	\hline
15	&	6.7	&	3.7	&	6.3	\\	\hline
16	&	10.4	&	7.2	&	9.1	\\	\hline
17	&	6.1	&	3.5	&	6.3	\\	\hline
18	&	\textbf{18.8}	&	\textbf{15.0}	&	\textbf{13.9}	\\	\hline
19	&	9.7	&	6.1	&	9.9	\\	\hline
20	&	\underline{5.0}	&	\underline{2.5}	&	6.0	\\	\hline
\hline
\end{tabular}
\end{center}
\end{table}
%
\subsection{Simulation: Planning Time}
\begin{figure} [b!]
\centerline{\includegraphics[width=3in]{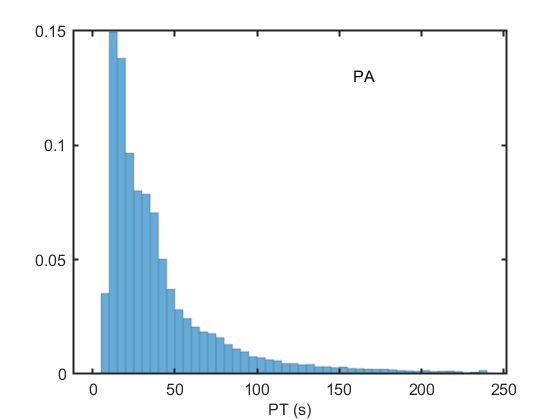}}
\caption{Normalized histogram of planning times (PT), in seconds for the entire simulation population (PA) of 30,000 trials. The population (PA) mean, median, and IQR are 53.7, 30, and 35 s.}
\label{fig:PT_PA}
\end{figure}

While TPL provides a measure of the success of MUONS in finding navigable paths, the time to plan the path is another important measure. Planning time (PT) is important not only because it should be included in the time to execute the mission, but it also allows operators to estimate the time to replan a path if the goal is changed or an obstacle is encountered. A normalized planning time histogram for the entire simulation population (PA) is provided in Figure \ref{fig:PT_PA}. For PA the mean, median and IQR of PT are 53.7, 30, and 35 seconds respectively. Planning times are measured in integer seconds with a minimum time of 5 seconds occurring 90 times in the simulation population (0.3\% of PA). The maximum reported planning time was 10587 seconds or nearly 3 hours. The value occurred for both the worst-performing mission (M4) and parameter set (PS18). PS18 also resulted in poor performance in terms of TPL.  Within PA, 2 samples are greater than 10,000 seconds also in M4, PS18. In fact, the MS4, PS18 subpopulation accounts for 35\% of times over 10 minutes (600 seconds). However, the total number of instances of these planning times is small in the total population at 142 (0.05\% of PA). Only 16 instances are an hour or more (15 of those in M4, P18). There are 14 instances where the planning time is between 30 minutes and one hour with 5 of those occurring within M4, P18. 

As with the other performance measures, we start by considering the effect of terrain first by examining cross sections of the population. When examining path length, we excluded M2 as an outlier creating the  PA$^{\prime}$ subpopulation, here we also excluded PS18. As we will see, in addition to being a poor choice with respect to TPL, it is also a clear outlier in PT. We refer to this subpopulation as PA$^{\prime \prime}$ (PA excluding M2 and P18) and note that the subpopulation mean, median, and IQR are 35.15, 27, and 28 s.  Considering both these statistics and Figure \ref{fig:PT_PA} we may also roughly approximate MUONS PT performance with an exponential distribution with a rate parameter of $1/\lambda_{PT.PA}$ = 30 s, though the actual time will depend upon specific implementation and compute platform.

Using PA$^{\prime \prime}$ as a reference population, we observe that the median planning time for P12 is 30 s and 24 s for P3, whereas for P2 median PT is 27 s. The median planning time for P1 and P1 excluding M2 is 35 s. If both M2 and M4 are excluded from P1 the median reduces further to 33.5 s. However, it is clear that the terrain instance (T1 to T2) has a more significant effect than terrain roughness alone on planning time increasing the median planning time by 24\%. 

\begin{figure} [b!]
\centerline{\includegraphics[width=3 in]{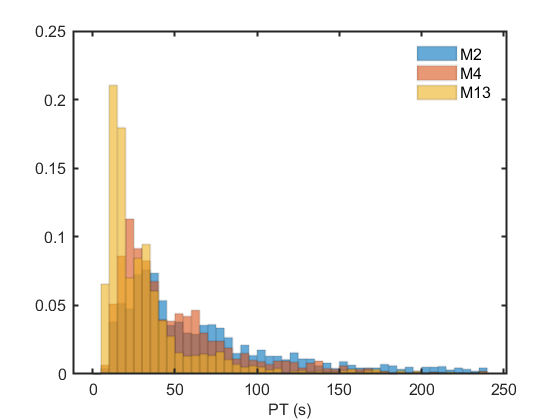}}
\caption{Normalized histogram of PT for M2, M4, and M13. M13 has the best performance in terms of planning time while M2 and M4 have the longest planning times.}
\label{fig:PT_M}
\end{figure}

We now turn our attention to the effect of mission on planning time performance. Summary statistics for PT by mission are found in Table \ref{tab:pt_m}. As discussed elsewhere, the M4 and M2 dominate the poor results with median planning times of 40 and 56~s respectively. It is interesting to note that while the statistics of M2 are worse than M4, M4 contains all the instances of planning times longer than 94 minutes, albeit a small number relative to the population. MUONS planning time for M13 was the shortest at 22 s with an IQR of 24 s. The minimum planning time was 5 s and 6.6\% of the M13 population was planned in 10 s or less. A similar percentage took 90 s or longer to plan with the longest taking 309 s. Normalized histograms of all three missions (M2, M4, M13) can be found in Figure \ref{fig:PT_M}. To summarize, depending on the choice of mission, the median planning time may be as much as 5 s shorter or 29 seconds longer. 
\begin{table}[h!]
\begin{center}
\caption{Planning time summary statistics by mission in seconds compared. The largest values across missions are in bold and the smallest are underlined.}
\label{tab:pt_m}
\begin{tabular}{l| c c c c c} 
\hline
\hline
Mission	&	Mean	&	Median	&	IQR	&	Min	&	Max	\\	\hline
1	&	51	&	34	&	38	&	7	&	371	\\	\hline
2	&	105	&	\textbf{56}	&	\textbf{73}	&	6	&	5646	\\	\hline
3	&	49	&	33	&	38	&	6	&	490	\\	\hline
4	&	\textbf{128}	&	40	&	47	&	7	&	10587	\\	\hline
5	&	50	&	34	&	37	&	5	&	518	\\	\hline
6	&	44	&	28	&	29	&	6	&	428	\\	\hline
7	&	40	&	27	&	28	&	6	&	390	\\	\hline
8	&	40	&	27	&	31	&	6	&	312	\\	\hline
9	&	39	&	25	&	27	&	5	&	502	\\	\hline
10	&	47	&	26.5	&	33	&	6	&	474	\\	\hline
11	&	68	&	30	&	32	&	5	&	2685	\\	\hline
12	&	\underline{34}	&	23	&	\underline{24}	&	5	&	265	\\	\hline
13	&	35	&	\underline{22}	&	\underline{24}	&	5	&	309	\\	\hline
14	&	38	&	23	&	25	&	5	&	2389	\\	\hline
15	&	38	&	23	&	25	&	5	&	706	\\	\hline
\hline
\end{tabular}
\end{center}
\end{table}

\begin{figure} [b!]
\centerline{\includegraphics[width=3in]{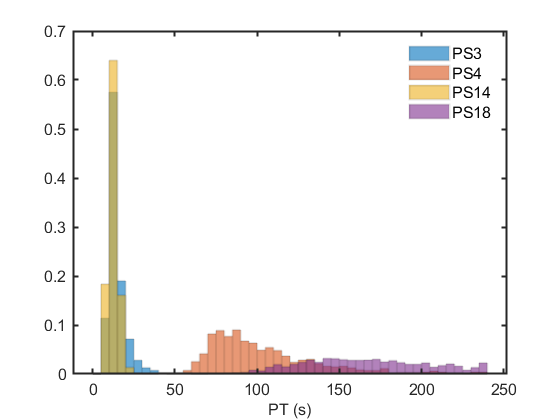}}
\caption{Normalized histogram of PT by PS3, PS4, PS14, and PS18. PS18 and PS4 have the worst overall performance in terms of PT and P14 the best. However, PS14 has the lowest SR among PS at 0.92. For this reason, PS3 is also included (SR 0.979).}
\label{fig:PT_PS}
\end{figure}

We have already alluded to the fact that parameter selection has a significant effect on PT. We now examine this relationship in more detail. A full account of the population statistics by each PS is provided in Table \ref{tab:pt_ps}. With a mean, median, and IQR of 356.9, 200, 151 s, respectively, PS18 is an obvious outlier taking 104 seconds longer to plan a path than the next worst set (PS4). These two PS account for all PTs greater than 10 minutes and 98.5\% of PTs greater than 5 minutes. Though, again, these account for a very small percentage of the population (PA).  PS14 has the overall best PT performance with a median planning time of 12 s and an IQR of 4 s. PS3 is comparable with PS14 with the same median but an IQR of 6 s. Normalized histograms of planning times for PS3, PS4, P14, and P18 are shown in Figure \ref{fig:PT_PS}. At this point, whether we consider TPL or PT, mission or parameter set, poor performance is characterized by a wide distribution that lacks the skew observed in the best-performing sections of the test population. Also, in the case of TP PS choice is a much larger factor than terrain or mission. Regarding that choice, most of the variability in PT is due to a poor choice of PS; three PS have the same median PT and similar IQRs. This suggests that it would be appropriate to choose a parameter set based on SR and TPL and then revise that selection if the PT is too long. 
\begin{table}[t!]
\begin{center}
\caption{PT summary statistics by PS in seconds compared. The largest values across missions are in bold and the smallest are underlined.}
\label{tab:pt_ps}
\begin{tabular}{l| c c c c c} 
\hline
\hline
PS	&	Mean	&	Median	&	IQR	&	Min	&	Max  \\ \hline
1	&	92	&	81	&	37	&	35	&	332 \\ \hline
2	&	21	&	19	&	9	&	5	&	73  \\ \hline
3	&	14	&	12	&	6	&	6	&	43  \\ \hline
4	&	110	&	96	&	43	&	37	&	979  \\ \hline
5	&	24	&	22	&	8.5	&	10	&	76  \\ \hline
6	&	20	&	18	&	8	&	9	&	52  \\ \hline
7	&	14	&	12	&	6	&	6	&	41  \\ \hline
8	&	30	&	28	&	22.5	&	6	&	108  \\ \hline
9	&	31	&	29	&	14	&	8	&	106  \\ \hline
10	&	43	&	41	&	39	&	5	&	276  \\ \hline
11	&	40	&	36	&	13	&	16	&	132  \\ \hline
12	&	45	&	40	&	16	&	22	&	154  \\ \hline
13	&	51	&	45	&	20	&	20	&	166  \\ \hline
14	&	\underline{12}	&	\underline{12}	&	\underline{4}	&	5	&	30  \\ \hline
15	&	24	&	21	&	9	&	9	&	79  \\ \hline
16	&	35	&	34	&	17	&	9	&	113  \\ \hline
17	&	17	&	15	&	6	&	7	&	50  \\ \hline
18	&	\textbf{357}	&	\textbf{200}	&	\textbf{151}	&	41	&	10587  \\ \hline
19	&	68	&	59	&	29	&	31	&	244  \\ \hline
20	&	27	&	25	&	10	&	13	&	81  \\ \hline
\hline
\end{tabular}
\end{center}
\end{table}

\subsection{Parameter Set Ranking}
Using Eq.\ref{eq:rank} we can rank both PS and MS. In the next section, we use these rankings to better understand the potential for further optimization at the mission level. These rankings are tabulated in the Appendix in Tables~\ref{tab:param_set_ranks} and~\ref{tab:mission_ranks}. Based on Eq.\ref{eq:rank} we choose PS3, 13, and 18 as a best, typical, and worst-performing set, respectively. Similarly, M13, M10, and M2 have these ranks across missions. 

\subsection{Mission-Based Parameter Tuning}
Now that we have evaluated the effect of terrain, mission, and parameter selection relative to our performance metrics, we aim to understand the potential benefits of further optimization. To accomplish this task, we look at the best, worst, and typical mission and assess the change in performance between the optimal PS for that mission and PS assessed as optimal over PA in terms of TPL via Eq.\ref{eq:rank}. 
\begin{figure}[hb!]
    \centering
z    \includegraphics[width=3in]{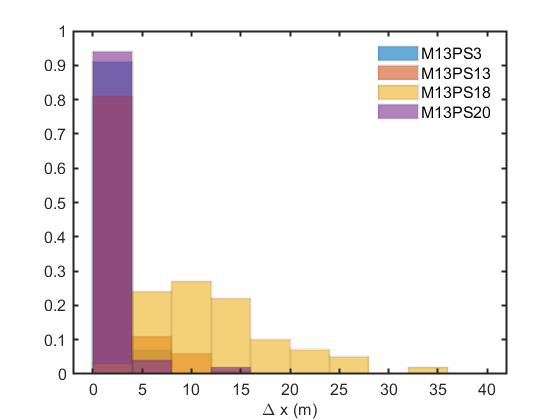}
    \caption{Normalized histogram of simulation excess TPL, $\Delta x$, across M13. MUONS planned the shortest paths for M13 in terms of $\Delta x$. PS3, PS13, and PS18 are the ``best'', ``typical'', and ``worst'' performing PS across the simulation population (PA). PS20 performs the best over this mission. However, as we have noted elsewhere PS20 has SR smaller than the mean over PA.}
    \label{fig:Best_TPL}
\end{figure}
\begin{figure}[hb!]
    \centering
    \includegraphics[width=3in]{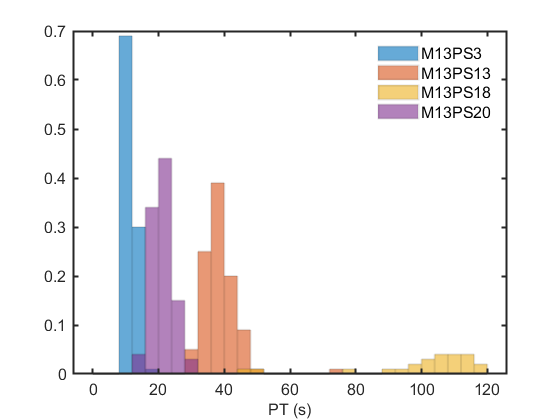}
    \caption{Normalized histogram of PT for M13 and PS3, PS13, PS18, and PS20. PS20 has the best TPL performance but is has a median PT that is 10 s slower. }
    \label{fig:Best_PT}
\end{figure}

We start with the mission where MUONS performed the best, M13, here median $\Delta x$ is 2.4 m and IQR 3.0 m. This mission is also the quickest to plan with a median PT of 22 s and IQR of 24 s. Within this mission applying P20 reduces the median $\Delta x$ to $0.92$ m and the IQR to 1.58 m. Though, as we remarked earlier, P20 has a relatively low SR (0.94). Our global best PS, PS3, results in $\Delta x$ of 1.7 m and an IQR of 2.1 m. The ``average’’ PS over PA, PS13, actually improves the median TPL to $\Delta x$ = 1.6 m with the same IQR. These results are clearly visualized in Figure \ref{fig:Best_TPL} where the poor performance of PS18 is also evident.  Both by inspection of Figure \ref{fig:Best_TPL} and the discussion above the improvement gained in TPL by further optimization; 0.7 m or less than the length of the AGV. The benefit of further optimization, when planner performance is already high, appears to be marginal. 

We have already remarked that PS20 has a relatively low SR, examining Figure \ref{fig:Best_PT} we also see that relative to PS3, the median planning time using PS20 is 10 seconds longer. For M13, PS3 provides the best overall PT. Using PS13 increases median PT to 37 s and PS18 to 158 s. In this case, the 0.7 m decrease in TPL is offset by an increase in PT of 10 s. 
\begin{figure}
    \centering
    \includegraphics[width=3in]{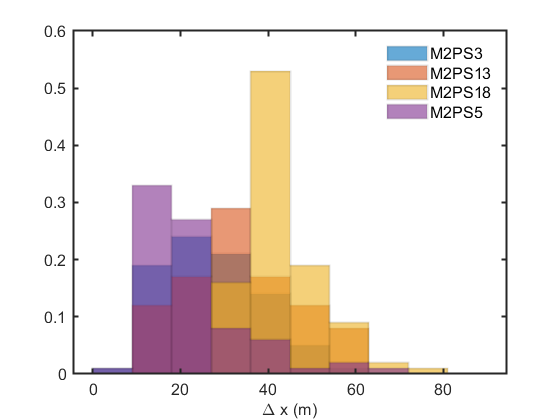}
    \caption{Normalized histogram of $\Delta x$ for PS3, PS13, PS18, and PS5 for M2. MUONS plans the longest and most inconsistent paths over M2. PS5 has the best TPL performance over M2. PS3, P13, and PS5 all have a median TPL smaller than the median over M2. In this case, PS18 has a smaller IQR at 10.1 m, though its median is 41.4 m}
    \label{fig:Worst_TPL}
\end{figure}
\begin{figure}
    \centering
    \includegraphics[width=3in]{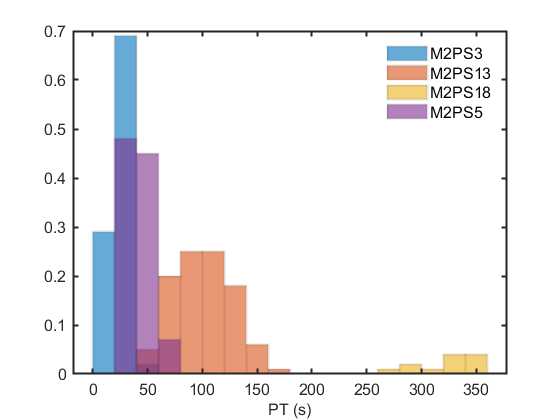}
    \caption{Normalized histogram of PT over M2 for PS3, PS13, PS18, and PS5. PS5 performs the best in terms of TPL but lags PS3 in terms of PT. Most PT values for PS18 are not visible in order to emphasize the results of the other PS.}
    \label{fig:Worst_PT}
\end{figure}

M2 has been noted several times so far as having particularly poor performance both in terms of TPL and PT.  It is natural to be curious the effect parameter selection has on this mission. Starting, again, with TPL, the sub-population median for M2 $\Delta x$ is 33.2 m and the IQR is 18.6 m. Over M2, PS5 has the smallest median $\Delta x$ at 19.4 m, though PS3 and PS13 also have medians smaller than the M2 sub-population at 26.6 and 31.5 m. For PS18 the median $\Delta x$ is 41.4 m, though as can be seen in Figure \ref{fig:Worst_TPL} where the IQR is relatively small at 10.1 m and comparable to PS5 at 10.5 m. It should not be surprising that PS18 takes the longest to plan a path at a median of 548 s with 144 s as the IQR; the shortest planning time for PS18 over M2 is 260 s. Comparing the normalized histograms for TPL and PT in Figures \ref{fig:Worst_TPL} and \ref{fig:Worst_PT}, the trade-off between TPL and PT becomes clearer. PS5 has the best TPL performance but follows behind PS3 in PT similar to what we observed in M13. 
\begin{figure}
    \centering
    \includegraphics[width=3in]{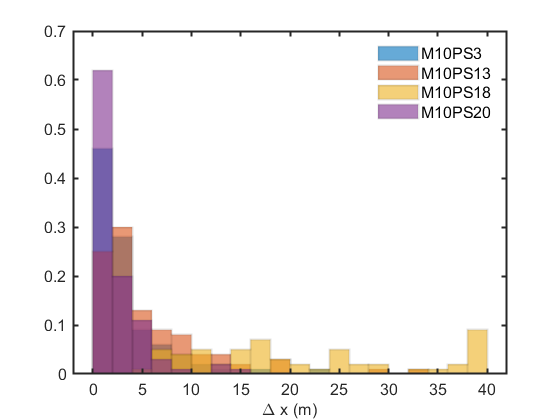}
    \caption{Normalized histogram of excess TPL, $\Delta x$, for PS3, PS13, PS18, and PS20 M10. MOUNS performance over M10 is typical of the simulation missions. PS20 plans slightly shorter paths than PS but at the expense of SR and PR.}
    \label{fig:AVG_TPL}
\end{figure}
\begin{figure}
    \centering
    \includegraphics[width=3in]{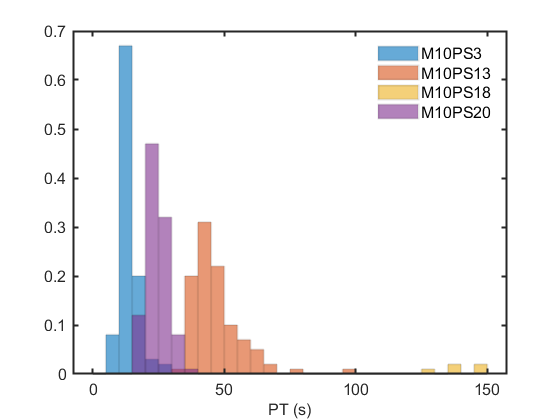}
    \caption{Normalized histogram of PT for M10 and PS3, PS13, PS18, and PS20. The median path planned by PS20 is 0.9 m shorter than PS3 but paths take 12 s longer to plan. The SR for PS20 is also lower than the median over PA.}
    \label{fig:AVG_PT}
\end{figure}

The worst and best missions (M2 and M13) are not typical. For this reason, it is prudent to also consider the effect of parameter tuning for a typical mission. We choose M10 for this purpose based on ranking with Eq.\ref{eq:rank}. The median $\Delta x$ over the M10 subpopulation was 4.2 m and the IQR 7.2 m. In terms of TPL for M10, applying PS20 results in the shortest median $\Delta x$ at 1.4 m and an IQR of 2.5 m. Using the PA highest ranked PS (PS3) increases the median by 0.9 m and the IQR by 0.7 m. If the average PS (PS13) is used median TPL increases to 3.7 m and the IQR to 5.6 m, larger than PS20 but smaller than the median and IQR of the M10 subpopulation. This relative performance is also observed in Figure \ref{fig:AVG_TPL} and we point out that median $\Delta x$ over M10, PS18 is 38.5 m. 

Over M10 the median planning time is 26.5 s with an IQR of 33 s, applying PS20 decreases the median planning time to 24 s and the IQR to 5 s. However, if PS3 is applied the median planning time is halved to 12 s and the IQR is reduced to 3.5 s.  Applying PS13 increases planning time relative to the subpopulation median to 44 s though the IQR is smaller at 10 s. The effect of PS18 on the population statistics is evident if one considers the minimum planning time for PS18 in M10 at 127 s compared to just 8 seconds for PS3. In Figure \ref{fig:AVG_PT} contributions from PS18 are present only in a few bins over the span between 0 and 150 s. 

The results presented in this section suggest against further optimization on a per-mission basis or even over terrain type (each of the selected missions were executed over different maps). When selecting the best-performing PS for each mission in terms of TPL the result was the same. While the best-performing PS decreased TPL slightly it also increased PT relative to the PS evaluated as best of the entire simulation population. 

\subsection{Mission and Parameter Set Discussion}
\begin{figure*}[ht!]
\begin{center}
\includegraphics[width=4in]{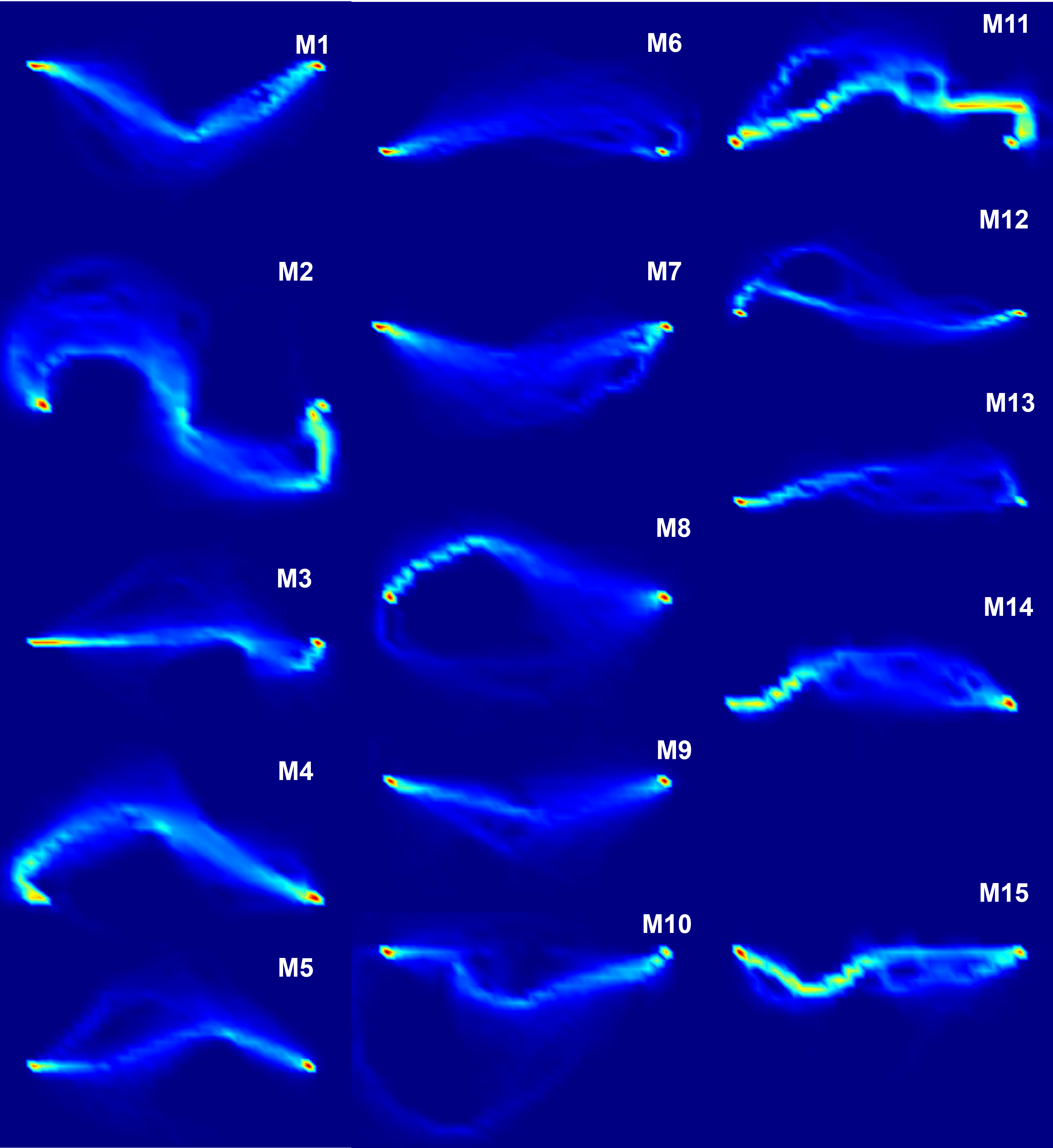}
\caption{Heatmaps of the planned path waypoints for each mission across all parameter sets. In each subfigure, the start position is located on the left and the goal at the right so that the straight-line path is along the horizontal. Each mission is labeled and missions are arranged in columns by terrain map (T1, T2, and T3).}
\label{fig:TPL_Mission_heatmap}    
\end{center}
\end{figure*}

In the preceding sections, we methodically assessed the variation in MUONS performance in terms of the mission tasks and planner parameter sets. Now that we have ranked both mission and parameter sets it makes sense to qualitatively examine what, if anything, differentiates these items. The uniqueness of M2 has been referenced repeatedly in our analysis. Figure \ref{fig:TPL_Mission_heatmap} includes heat maps of the planned paths for each mission. In each subfigure, the start is oriented on the left and the goal on the right so that the straight-line path is along the horizontal. Columns are arranged by terrain type (T1, T2, and T3) for ease of comparison. From even a cursory glance at this Figure it is clear that M2 is an outlier. Several features make this heatmap unique. First, the initial path from the start position is directed away from the goal. The "S" shape of the planned paths is another unique feature. From our experience, this configuration is particularly difficult for Bi-RRT to plan. While MUONS includes heuristics for joining quickly such paths, it is possible that additional improvements could be made in addressing this condition. Relative to causal factors, examining Figure \ref{fig:T1-3_rand_missions} we observe that the straight-line path for M2 starts behind a steep terrain feature and traverses a depression, making this mission particularly difficult to plan. Inspecting M4 in Figure \ref{fig:T1-3_rand_missions}, we observe it shares many features with M2 and, accordingly, has similar planning times. 

M13 was selected where MUONS was most successful in planning a path. Examining Figure \ref{fig:T1-3_rand_missions} we observe that M13 is a mostly straight-line path so this is not unexpected. M10 was selected as the typical mission though comparison with the other missions in Figure \ref{fig:TPL_Mission_heatmap} shows it to be more atypical. In our assessment, a typical path morphology is that of an arc or single deviation around a challenging terrain segment. While M10 does include these features it also includes at least one planned path traversing a long arc along the map boundary. A close inspection of our results shows that these paths are associated with PS18. Referring again to Figure \ref{fig:T1-3_rand_missions}, the straight-line path for M13 is mostly over relatively flat terrain, while M10 has what may be interpreted as a tight corridor path. 

Moving on to our parameter sets we start with PS18, this parameter set is remarkable in that it has the smallest value of $N^{avg}_{max}$ at 8.23. However, PS10 and PS14 performed with similar values at $N^{avg}_{max}$ of 15.4 and 14.64. PS18 also has the smallest value of $r^{bi-rrt}_{exp}$ at 0.24 and we observe that PS1 and PS19 also perform poorly with Bi-RRT expansion radii of 0.5 and 0.65 respectively. As we will see in the next section, this parameter has the highest correlation with our performance metrics. 

Assessing what differentiates the best-performing PS from the group is more difficult. PS3 has a larger $r^{bi-rrt}_{exp}$ at 1.84 and weighting parameter $\alpha$. PS14 has the largest $r^{bi-rrt}_{exp}$ (2.83) and was observed to provide the best planning time result. PS20, a frequent mission optimum has a slightly larger value of $\alpha$ though was noted as having a lower SR. As one might expect, these results suggest that the optimum PS balances between parameters. It also suggests that further improvements may be possible via iterative optimization methods. 
\subsection{Correlation Analysis}
\label{sec:SA_sim}
\begin{figure} [hb!]
\centerline{\includegraphics[width=3.5in]{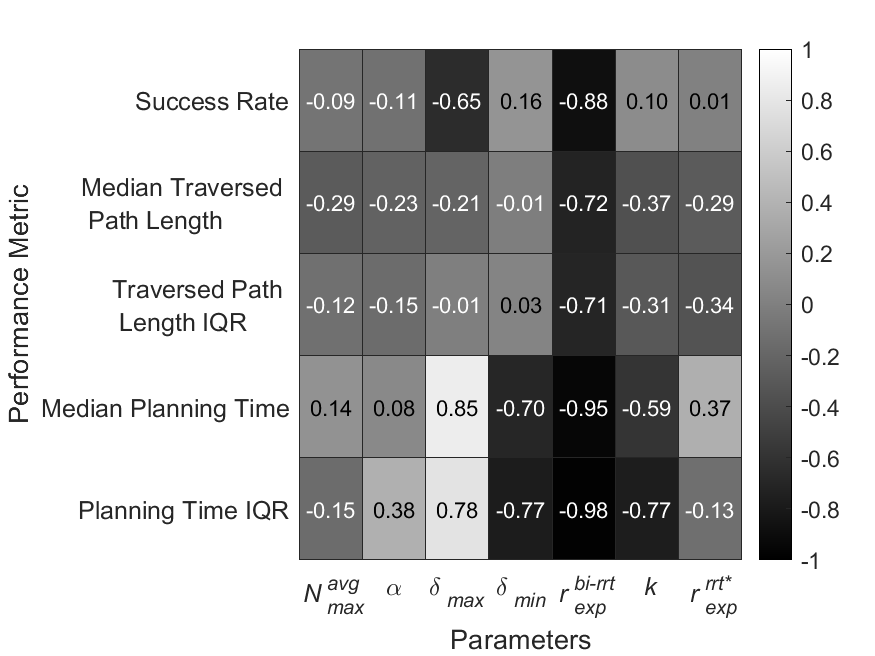}}
\caption{Spearman partial correlation coefficients between the seven path planning parameters and the success rate, and the median and IQR metrics for planning time and traversed path length.}
\label{fig:SA_spearman_agg}
\end{figure}

Our discussion in the previous section alludes to strong correlations between performance metrics and the sampled values of the path-planning parameters. A heatmap of the Spearman partial correlation coefficients illustrating these correlations is shown in Fig.~\ref{fig:SA_spearman_agg}. The heatmap shows that $r^{bi-rrt}_{exp}$ exhibits the strongest correlation with each performance metric ($r_{partial} \leq -0.71$) compared to the other parameters. The strong negative correlation with the success rate is due to increased waypoint spacing, which raises the likelihood of encountering untraversable terrain between waypoints, leading to more failures. The algorithm assumes waypoints are proximate and does not evaluate inter-waypoint traversability.

The $r^{bi-rrt}_{exp}$ column also shows a strong negative correlation with the median and IQR of TPL. Increased waypoint spacing results in longer path segments, which leads to shorter overall planned and traversed path lengths on average, with less variability.

At the bottom of the $r^{bi-rrt}_{exp}$ column, there is a strong negative correlation with the median and IQR of planning time. Larger waypoint spacing allows the Bi-RRT algorithm to expand faster, finding initial paths more quickly and with less variability. Larger waypoint space also results in fewer waypoints, requiring less processing in downstream planning steps. The significant influence of $r^{bi-rrt}_{exp}$ on performance is due to its role in the initial path found with Bi-RRT, which is then used in subsequent path-planning phases.

In terms of secondary effects, the parameters associated with LTO, $\delta_{min}$ and $\delta_{max}$ are observed to be well correlated with PT (0.7 to 0.85) and symmetrically so. This suggests that as $\delta_{min}$ is increases median and IQR of PT decreases. Though we note that variation of these two parameters is related as noted in Table \ref{tab:param_dist}. Interestingly, $\delta_{max}$ is negatively correlated with SR.  The start-to-goal RRT* proportional constant, $k$, is also negatively correlated with PT. The correlation results also indicate a tradeoff in performance: as $r^{bi-rrt}_{exp}$ increases, the aggregated success rate decreases along with the median and IQR of planning times and traversed path lengths. This correlation analysis identifies $r^{bi-rrt}_{exp}$ as a crucial parameter for tuning to achieve reliable navigation while minimizing planning times and path lengths. 

\section{Field Experiments}
\label{sec:field_results}
%
\begin{figure*}[tb!]
\centering
\subfigure[]{\label{fig:FT1-2-rock2-annotated}\includegraphics[width=2.35in]{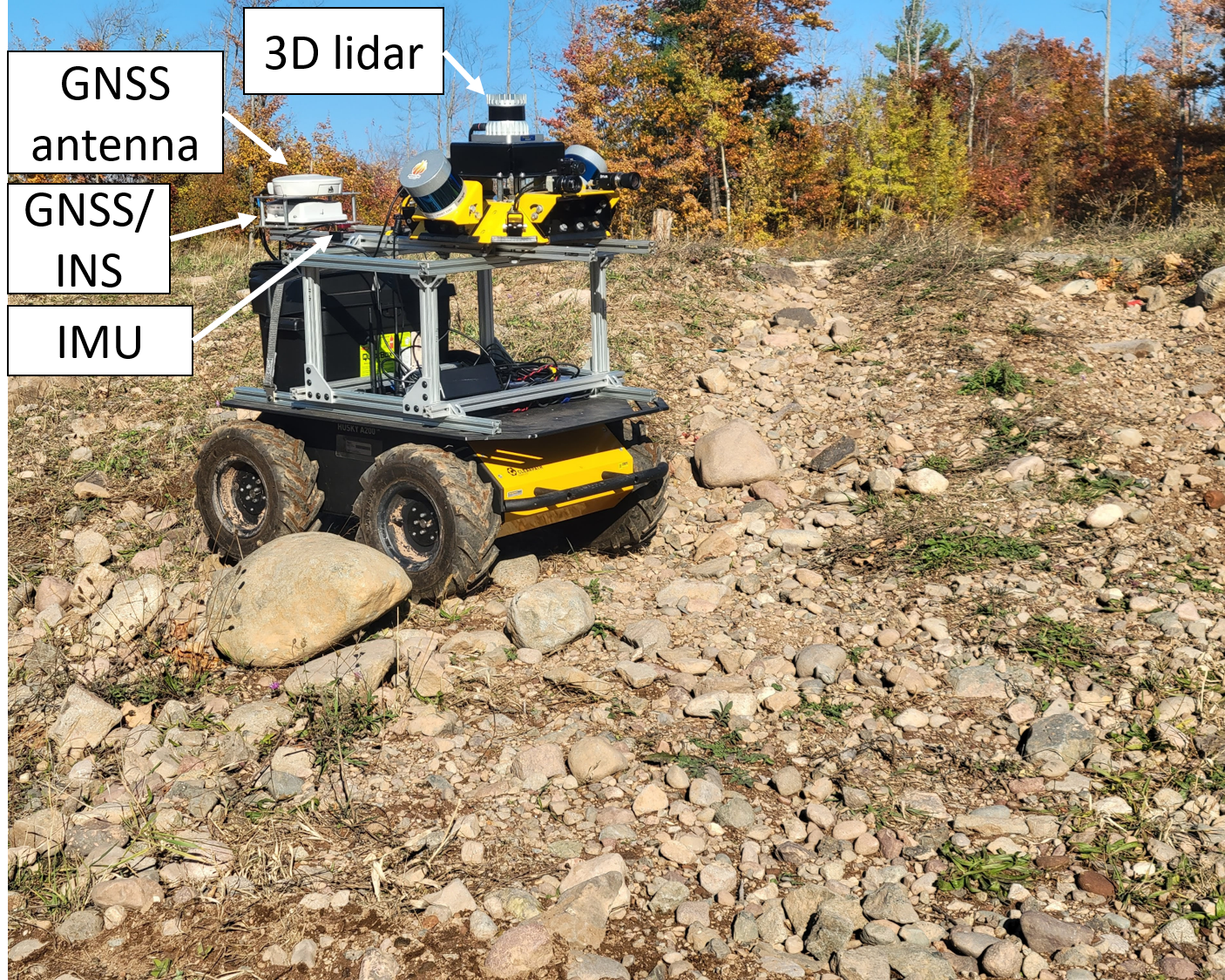}}
\subfigure[]{\label{fig:}\includegraphics[width=2.35in]{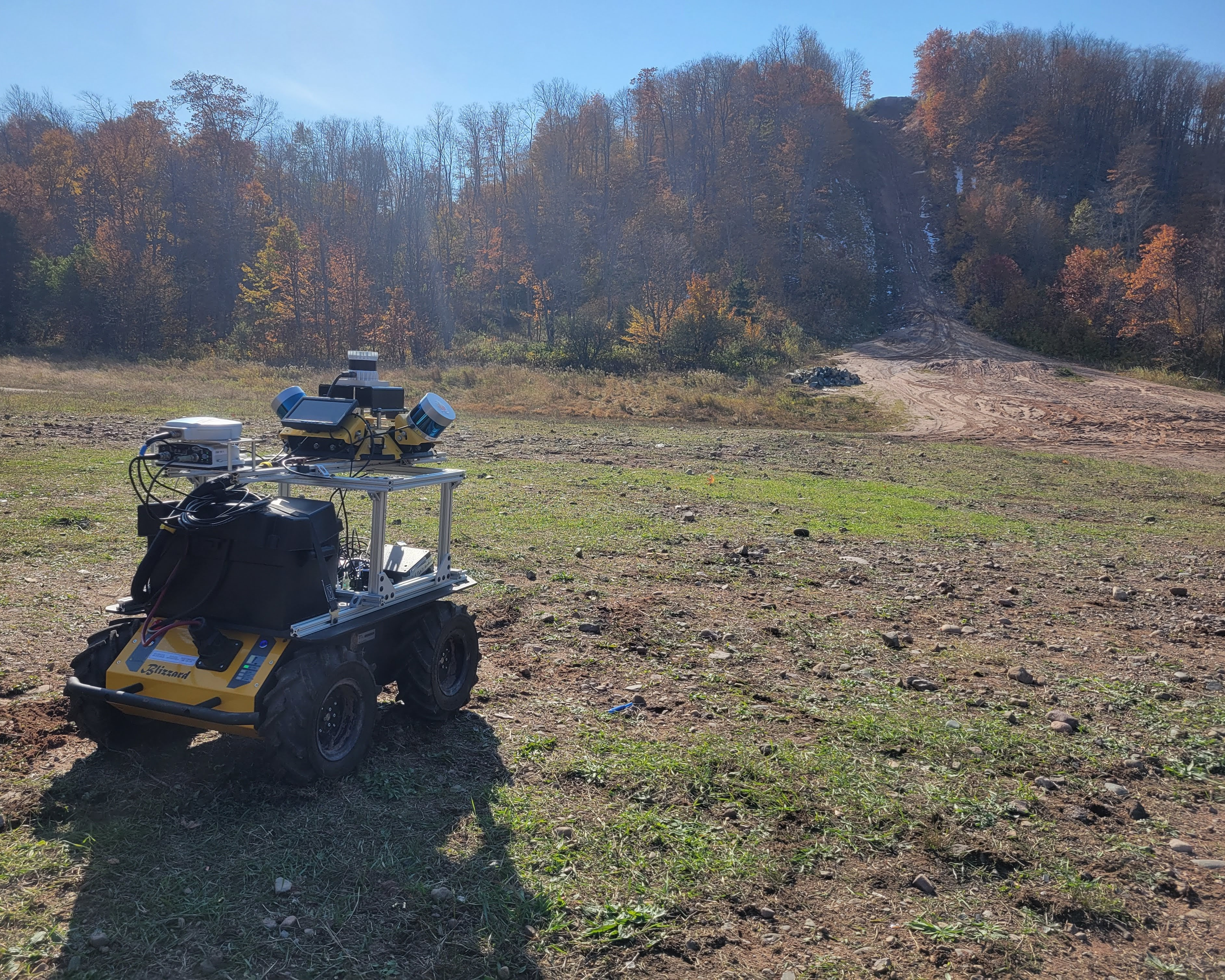}}
\subfigure[]{\label{fig:FT2_start}\includegraphics[width=2.35in]{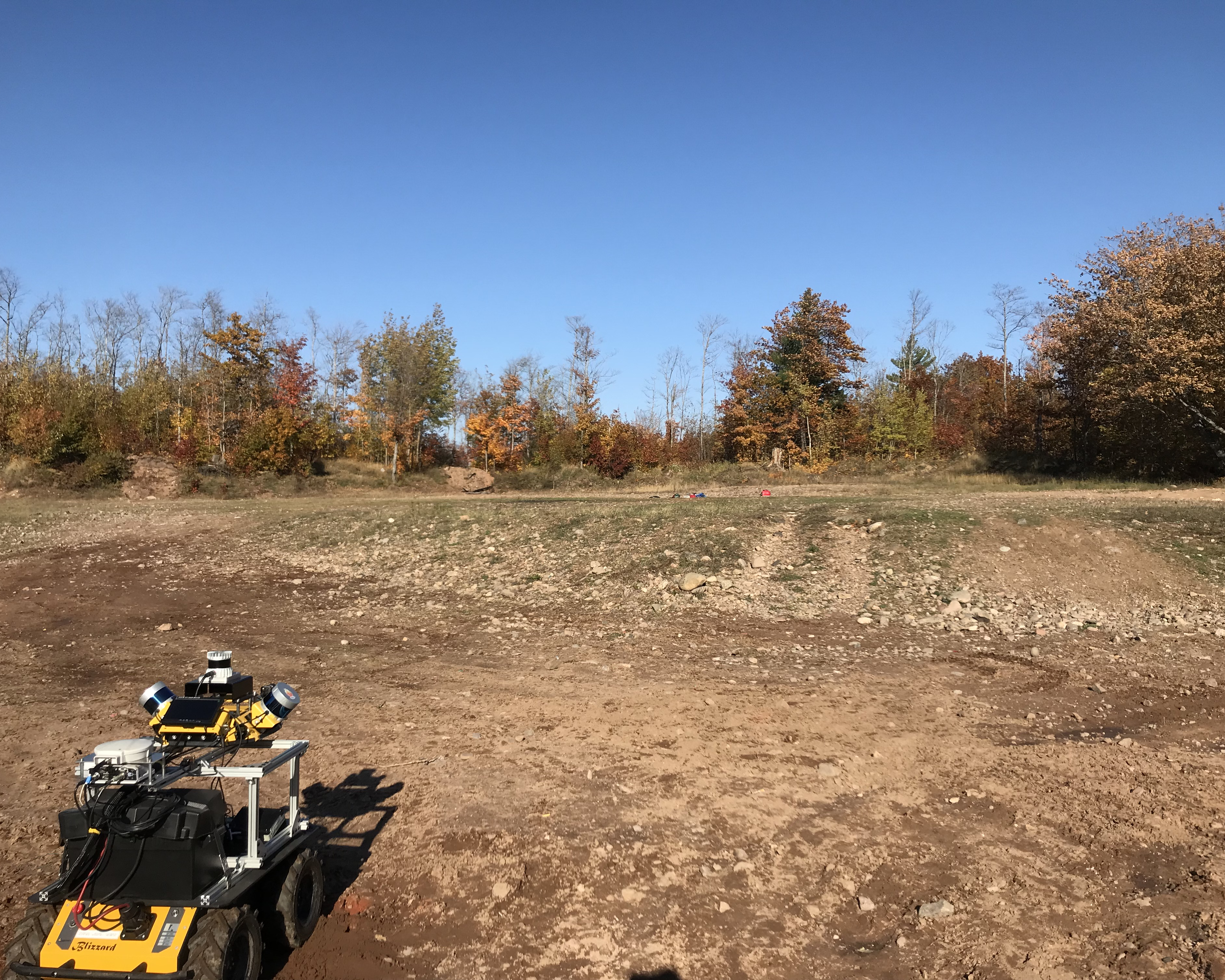}}
\subfigure[]{\label{fig:SR_op_area}\includegraphics[width=2.35in]{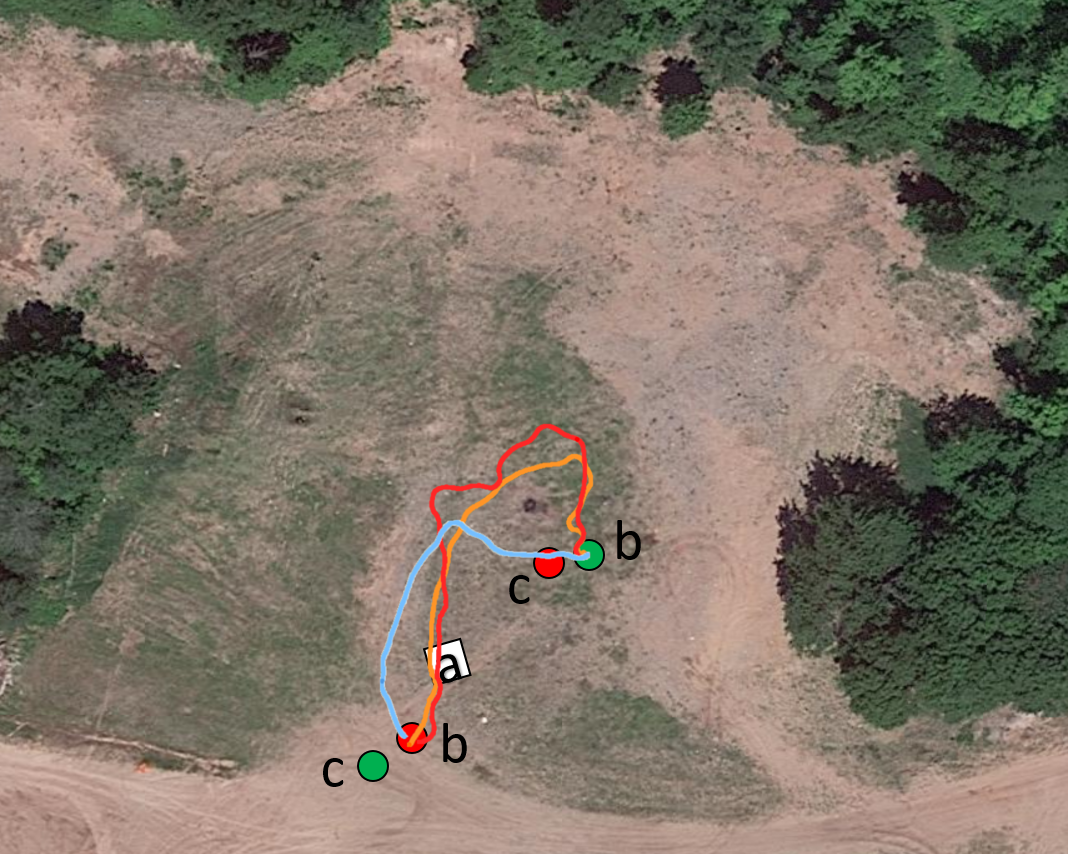}}
\caption{Outdoor navigation experiments in rough, off-road terrain. The Husky AGV is positioned between the start and goal for FTs 1 and 2  (a). In (b) the AGV is shown at the start pose for FT1. The start pose for FT2 is pictured in (c). A satellite image of the test area~\protect\cite{GoogleMaps} is shown in (d) with the start and goal locations in FTs 1 and 2 shown as green and red dots labeled with "b" and "c", respectively. Example paths for FT1 using parameter sets 3 (blue), 13 (orange), and 18 (red) are also shown in (d). The white rectangle in (d) reflects the AGV's location in (a).}
\label{fig:FT_all_figs}
\end{figure*}

In Section~\ref{sec:sim_results}, we analyzed the performance of MUONS in simulation and selected PS3, P13, and P18 as ``best’’, ``typical’’, and ``worst’’ performing. In this section, we lay out a plan for field testing where the relative performance of these PS are evaluated in the field. In addition to spanning the range of performance observed in the simulation trials, the selected PS also demonstrate high success rates (above $98\%$), minimizing the risk of hardware damage during testing. 

\subsection{Robotic Platform}
\label{sub:hardware}
For our field experiments, we use a Clearpath Husky A200, the same robotic platform modeled in simulation. This four-wheeled, skid-steered robot is equipped with several sensors, with the relevant ones annotated in Fig. 7(a). A NovAtel Pwrpak7-E1 global navigation satellite system (GNSS) with an integrated inertial navigation system (INS) and Real-Time Kinematic (RTK) corrections is used for localization. An Ouster OS1-64 LiDAR sensor and a VectorNav VN-100 inertial measurement unit (IMU) allow the generation of the point-cloud map used for path planning as described in Section~\ref{sec:local_and_map}.

\subsection{Field Testing Overview}
MUONS is tasked with navigating over off-road terrain containing rocks, gravel, and grass patches. All field tests (FTs) feature sloped terrain between the start and goal poses, making a straight-line path unlikely due to enforced maximum pitch and roll constraints. Figures~\ref{fig:FT1-2-rock2-annotated} - \ref{fig:FT2_start} show the Husky at various locations in the test environment.

The FTs consist of 30 trial runs across two different start-to-goal tasks, using parameter sets 3, 13, and 18. These tasks are referred to as \emph{FT1} and \emph{FT2}. An aerial view of the operational area, annotated with the respective FT start/goal locations, is shown in Fig.\ref{fig:SR_op_area}, along with an example path traversed by the AGV for each parameter set. Each navigation task's start and goal pose are separated by a Euclidean distance of 34m. Each parameter set was tested five times per task. To reduce the risk of damaging the AGV, the maximum allowable pitch and roll constraints were set to $(-0.15, 0.15)$ radians.

\subsection{Field Testing Results and Comparison to Simulation}
\label{sec:FT_SR}
In the field, the AGV completed all 30 navigation trials. This success mirrors the reliability observed in completing over 4,410 of the 4,500 trials in simulation using parameter sets 3, 13, and 18. The median time to plan a path (PT) was about 3 s compared to 35 s in our simulation experiments. The excess path length, $\Delta x$, was 38.8 m over FT1 and FT2 compared to 4.6 m over PA. With only 30 samples and considering the variation introduced by both mission and PS, second-order statistics are less meaningful but will be included in tabulated results for completeness. Full tabulated results for PT and TPL are found in Tables~\ref{tab:field_raw_PT} and \ref{tab:field_raw_TPL} respectively in the appendix. 

In Table~\ref{tab:sim_field_PT_summary} we compare the summary statistics of PT between simulation and field testing. Note that these statistics are computed over around 1,500 samples for simulation trials and just 10 for field testing. The potential variables here are the compute platform used for planning and the size of the planning map. For this reason, we expect good agreement and that is what we observe. Between PS3 and PS13 in simulation, the ratio of the median PT (PS3/P13) is 0.27 in simulation and 0.28 for field testing. A similar relationship can be found between the other statistics presented. Comparing PS18 to PS3 the ratio of median planning times (PS18/PS13) is 4.44 in simulation and 5.46 for field testing. Considering an IQR of 151 seconds was observed in our simulation testing, this is a favorable result for only 10 samples. 
\begin{table}[b!]
\begin{center}
\caption{Summary statistics comparing our simulation and field-testing results in terms of PT. All results are in seconds. }
\label{tab:sim_field_PT_summary}
\begin{tabular}{l| c c c c c c} 
\hline
\hline
Metric	&	 \multicolumn{6}{c}{\underline{\hspace{72pt} PS \hspace{72pt}}}	\\
	&	\multicolumn{3}{c}{\underline{\hspace{21pt} Simulation \hspace{21pt}}}	&	\multicolumn{3}{c}{\underline{\hspace{22pt} Field \hspace{22pt}}} \\ 
	&	3	&	13	&	18	&	3	&	13	&	18	\\	\hline
Mean	&	13.9	&	50.9	&	356.9	&	1.0	&	3.5	&	19.5	\\	\hline
Median	&	12.0	&	45.0	&	200.0	&	0.8	&	2.9	&	16.0	\\	\hline
IQR	&	6.0	&	20.0	&	151.0	&	0.3	&	1.9	&	10.9	\\	\hline
$\sigma$	&	5.2	&	19.5	&	709.2	&	0.5	&	1.3	&	8.5	\\	\hline
\hline
\end{tabular}
\end{center}
\end{table}

In Table~\ref{tab:sim_field_TPL_summary} the summary statistics of $\Delta x$ are compared between simulation and field trials. Comparing the results here is more complicated owing to how path length is recorded in both experiments. In simulation, this quantity can be recorded exactly, whereas in the field we rely on MUONS localization via an EKF as ground truth. The variability in our simulation results for PS18 is another confounding factor. Considering all these factors the relative agreement between simulation and field trials is remarkable. In simulation, the ratio of median $\Delta $ between PS3 and PS13 is 0.66 compared to 0.60 for our field trials. The other summary statistics, though perhaps only notional, are similar. Comparing PS18 to PS13 the ratio of median excess path length (PS18/PS13) is 3.55 in simulation and 1.76 in field trials. It is interesting to note that the ratio of the standard deviations between these PS is around 2 in both simulation and our field data. 

\begin{table}[b!]
\begin{center}
\caption{Summary statistics comparing our simulation and field-testing results for TPL in terms of $\Delta x$. All results are in meters.}
\label{tab:sim_field_TPL_summary}
\begin{tabular}{l| c c c c c c} 
\hline
\hline
Metric	&	 \multicolumn{6}{c}{\underline{\hspace{68pt} PS \hspace{68pt}}}	\\
	&	\multicolumn{3}{c}{\underline{\hspace{10pt} Simulation \hspace{10pt}}}	&	\multicolumn{3}{c}{\underline{\hspace{28pt} Field \hspace{28pt}}} \\ 
	&	3	&	13	&	18	&	3	&	13	&	18	\\	\hline
Mean	&	4.6	&	6.2	&	18.7	&	26.1	&	31.3	&	60.4	\\	\hline
Median	&	2.8	&	4.2	&	15.0	&	17.7	&	29.6	&	52.0	\\	\hline
IQR	&	4.9	&	7.4	&	13.9	&	24.1	&	17.5	&	18.4	\\	\hline
$\sigma$	&	5.5	&	5.9	&	12.8	&	17.2	&	14.4	&	29.6	\\	\hline
\hline
\end{tabular}
\end{center}
\end{table}

\section{Conclusions and Future Work} 
\label{sec:conclusions}
In this paper, we have successfully demonstrated that planning on point-cloud maps is a robust method for planning paths over kinematically challenging terrain. We establish robustness by evaluating MUONS in both an extensive simulation campaign featuring 30,000 unique "A to B" traversals and a limited field campaign. Over these campaigns, we used success rate, traversed path length, and planning time as performance metrics. 

In simulation, we demonstrated that MUONS improves success rate by between 0.23 and 0.47 depending on the complexity of the terrain compared to following the straight-line path. In over 30,000 trials the simulated AGV failed to complete the task in just over 600 cases for a combined SR of 0.98 over three different terrain maps. After accounting for outliers, the effect of terrain difficulty on SR when applying MUONS was measured to be only 0.01 compared to 0.25 following a straight-line path. In aggregate, it was shown that the choice of planning parameters can reduce SR by 0.04 (0.94). For missions that are difficult to plan (i.e. M2) a poor choice of MUONS PS results in a SR that is no worse than the straight-line results. 

Of course, the increase in success rate is not without cost. We found that the increase in path length, $\Delta x$, required to achieve the high SR can be well-approximated by an exponential distribution with a mean of 7 m for a 35 m straight-line path. Equivalently, the mean $\Delta x$ introduced by MUONS is about 20\% of the path length. The degree to which $\Delta x$ scales with path length for a given terrain is unclear and a possible direction for future work. At the same time, we expect that this result will be useful to operators and researchers seeking to quantify uncertainty in off-road autonomy. A similar result was observed in assessing MUONS planning time, though this quantity is dependent on map size and compute platform. The mean planning time for our simulations was 37.7 s excluding PS18 but as low as 1 second in our field trials. 

Due in part to the observation that path and planning penalties appear exponentially distributed, there is a case to be made that MUONS performance is invariant to terrain roughness. This is supported by comparing results across terrain maps T2 and T3. On the other hand, T1 which has the same fractal statistics as T2 contained several missions that proved difficult to plan and supports the conclusion that choice of mission accounts for more variability in performance than terrain or choice of path planning parameters. Assuming, of course, that the PS is not an outlier like PS18. We observed that choice of mission can increase median excess path length by as much as 50\% and variability by 14\% compared to our modified simulation population. 

Relative to the choice of path planning parameters, we observed the choice of PS can significantly affect TPL but mostly in the negative. When using the best performing PS the median excess path length was 3 m compared to 15 m for the worst compared to a median over the simulation population of 4.6 m. The trend is similar when considering the choice of PS on PT, but the effect is much greater. The best median planning time across PS was 12 s and the worst 200 s compared 27 s for our sample population. 

To understand the potential benefits of tuning path parameters on a per-mission basis we ranked missions across terrain types. The same method was used to establish a best, typical, and worst-performing PS as PS3, PS13, and PS18 respectively. We examined the difference between the median $\Delta x$ and PT for PS that performed the best in each mission compared to those selected over the simulation population. We found that improvements on a per-mission basis invariably involved a reduction in SR or an increase in PT. For this reason, a major finding of this work is that parameter tuning is likely best done for a specific platform over a variety of terrains and missions. Optimizing a planner over a specific mission is likely to not generalize to other missions even on the same terrain. 

Correlation analysis revealed that the Bi-RRT expansion radius has the most significant impact on performance across all performance metrics. Though, qualitatively this effect is mostly negative; a proper choice for this parameter is crucial. However, it is unclear how much performance is to be gained from finding the optimal rather than a merely adequate value. There are clear trade-offs as well considering that PS with a $r^{bi-rrt}_{exp}$ larger than about $1,8$ have a lower SR. 

To validate our simulation results we conducted a limited field-testing campaign with the aim of comparing performance between PS3, PS13, and PS18. We found that the proportional variation in the first-order statistics (mean and median) over PS matched exceptionally between simulation and the field. While simulation was not predictive of mission performance in the field the approach and results described in this paper support the use of modeling and simulation for performance tuning. Specifically, our findings suggest that a simulation environment where mission and terrain are randomly generated may allow for tuning of planner parameters using straightforward optimization methods or machine learning methods which is a direction of future work. 

\section*{Acknowledgments}
The authors wish to acknowledge the financial support provided by the USD/R\&E (The Under Secretary of Defense-Research and Engineering), National Defense Education Program (NDEP)/BA-1, Basic Research in the Science Mathematics and Research for Transformation (SMART) Scholarship Program. Technical and financial support was also provided by the Automotive Research Center (ARC) under Cooperative Agreement W56HZV-19-2-0001 U.S Army CCDC Ground Vehicle Systems Center (GVSC) Warren, MI. These views and/or findings do not represent the views of the SMART Program, Department of Defense, United States Government, ARC, or GVSC. (Corresponding author: Casey Majhor.) We would also like to acknowledge contributions by Sam Kysar, Parker Young, Zach Jeffries, Ian Mattson, and Jake Carter.


\section*{Conflict of interest}
The authors declare no conflict of interest.

\section*{Supporting information}
Additional supporting information may be found in the online version of the article on the publisher’s website.




%
% ...

\bibliographystyle{asmems4}

\bibliography{asme2e}

\appendix{}

\begin{table}[!th]
\caption{Parameter sets and their corresponding seven path-planning parameter values.}
\label{tab:param_sets}
\centering
\resizebox{\columnwidth}{!}{%
\begin{tabular}{l c c c c c c c} 
\hline\hline
Parameter Set & $r^{bi-rrt}_{exp}$ & $k$ & $r^{rrt*}_{exp}$ & $\alpha$ & $N^{avg}_{max}$ & $\delta_{max}$ & $\delta_{min}$ \\
\hline
1	&	0.50	&	0.14	&	0.18	&	0.19	&	43.57	&	0.57	&	0.09	\\	\hline
2	&	1.56	&	0.02	&	0.08	&	0.30	&	34.21	&	0.22	&	0.06	\\	\hline
3	&	1.84	&	0.21	&	0.09	&	0.76	&	28.40	&	0.18	&	0.03	\\	\hline
4	&	0.51	&	0.09	&	0.04	&	0.03	&	25.09	&	0.96	&	0.08	\\	\hline
5	&	1.94	&	0.11	&	0.12	&	0.59	&	43.03	&	0.77	&	0.06	\\	\hline
6	&	2.56	&	0.11	&	0.07	&	0.20	&	64.48	&	0.29	&	0.03	\\	\hline
7	&	2.38	&	0.05	&	0.04	&	0.47	&	31.84	&	0.30	&	0.10	\\	\hline
8	&	0.82	&	0.08	&	0.05	&	0.26	&	67.24	&	0.24	&	0.05	\\	\hline
9	&	0.97	&	0.09	&	0.16	&	0.12	&	85.46	&	0.22	&	0.04	\\	\hline
10	&	0.58	&	0.06	&	0.15	&	0.19	&	15.40	&	0.15	&	0.04	\\	\hline
11	&	1.37	&	0.08	&	0.21	&	0.49	&	77.92	&	0.80	&	0.08	\\	\hline
12	&	1.01	&	0.12	&	0.14	&	0.57	&	69.86	&	0.91	&	0.06	\\	\hline
13	&	0.87	&	0.12	&	0.09	&	0.49	&	52.47	&	0.83	&	0.07	\\	\hline
14	&	2.83	&	0.04	&	0.01	&	0.11	&	98.29	&	0.05	&	0.02	\\	\hline
15	&	1.23	&	0.14	&	0.13	&	0.07	&	14.64	&	0.24	&	0.03	\\	\hline
16	&	1.17	&	0.02	&	0.07	&	0.35	&	64.91	&	0.18	&	0.02	\\	\hline
17	&	1.81	&	0.13	&	0.12	&	0.38	&	28.36	&	0.29	&	0.10	\\	\hline
18	&	0.24	&	0.16	&	0.19	&	0.38	&	8.23	&	0.36	&	0.03	\\	\hline
19	&	0.65	&	0.26	&	0.05	&	0.26	&	62.99	&	0.41	&	0.03	\\	\hline
20	&	2.20	&	0.11	&	0.05	&	0.77	&	89.22	&	0.77	&	0.04	\\	
\hline\hline
\end{tabular}%
}
\end{table}

\begin{table}[th!]
\begin{center}
\caption{Parameter set ranks based on Eq.\ref{eq:rank}.}
\label{tab:param_set_ranks}
\begin{tabular}{l| c} 
\hline
\hline
Parameter Set	&	Rank	\\	\hline
1	&	0.41	\\	\hline
2	&	0.55	\\	\hline
3	&	0.72	\\	\hline
4	&	0.47	\\	\hline
5	&	0.72	\\	\hline
6	&	0.69	\\	\hline
7	&	0.59	\\	\hline
8	&	0.64	\\	\hline
9	&	0.70	\\	\hline
10	&	0.50	\\	\hline
11	&	0.69	\\	\hline
12	&	0.67	\\	\hline
13	&	0.60	\\	\hline
14	&	0.50	\\	\hline
15	&	0.71	\\	\hline
16	&	0.52	\\	\hline
17	&	0.70	\\	\hline
18	&	0.17	\\	\hline
19	&	0.54	\\	\hline
20	&	0.70	\\	\hline
\hline
\end{tabular}
\end{center}
\end{table}

\begin{table}[th!]
\begin{center}
\caption{Mission ranks based on Eq.\ref{eq:rank}.}
\label{tab:mission_ranks}
\begin{tabular}{l| c} 
\hline
\hline
Mission	&	Rank	\\	\hline
1	&	0.64	\\	\hline
2	&	-0.56	\\	\hline
3	&	0.71	\\	\hline
4	&	0.42	\\	\hline
5	&	0.74	\\	\hline
6	&	0.71	\\	\hline
7	&	0.66	\\	\hline
8	&	0.53	\\	\hline
9	&	0.81	\\	\hline
10	&	0.67	\\	\hline
11	&	0.35	\\	\hline
12	&	0.65	\\	\hline
13	&	0.84	\\	\hline
14	&	0.83	\\	\hline
15	&	0.62	\\	\hline
\hline
\end{tabular}
\end{center}
\end{table}

\begin{table}[h!]
\begin{center}
\caption{Field PT results in seconds}
\label{tab:field_raw_PT}
\begin{tabular}{l| c c c c} 
\hline
\hline
FT    &   Trial   & \multicolumn{3}{c}{\underline{\hspace{22pt} PS \hspace{22pt}}} \\
	&		&	3	&	13	&	18	     \\	\hline
	&	1	&	0.9	&	2.9	&	13.5	\\	
	&	2	&	1.0	&	3.0	&	29.4	\\	
FT1	&	3	&	0.8	&	2.4	&	34.6	\\	
	&	4	&	1.0	&	5.7	&	15.6	\\	
	&	5	&	0.6	&	2.4	&	29.6	\\	\hline
	&	1	&	0.7	&	3.1	&	17.1	\\	
	&	2	&	0.7	&	2.7	&	15.4	\\	
FT2	&	3	&	1.1	&	2.8	&	8.2	     \\	
	&	4	&	0.8	&	5.1	&	15.5	\\	
	&	5	&	2.2	&	5.3	&	16.3	\\	\hline
\hline
\end{tabular}
\end{center}
\end{table}

\begin{table}[th!]
\begin{center}
\caption{Field TPL results in meters}
\label{tab:field_raw_TPL}
\begin{tabular}{l| c c c c} 
\hline
\hline
FT    &   Trial   & \multicolumn{3}{c}{\underline{\hspace{28pt} PS \hspace{28pt}}} \\
	&		&	3	&	13	&	18	     \\	\hline
	&	1	&	16.9	&	31.1	&	44.1	\\	
	&	2	&	59.7	&	22.3	&	91.2	\\	
FT1	&	3	&	16.3	&	23.8	&	128.0	\\	
	&	4	&	45.0	&	60.3	&	63.7	\\	
	&	5	&	13.7	&	28.2	&	61.5	\\	\hline
	&	1	&	18.4	&	42.1	&	51.6	\\	
	&	2	&	8.5	&	11.4	&	45.1	\\	
FT2	&	3	&	24.0	&	16.3	&	22.1	\\	
	&	4	&	14.4	&	41.4	&	44.6	\\	
	&	5	&	44.1	&	36.3	&	52.5	\\	\hline

\hline
\end{tabular}
\end{center}
\end{table}

\end{document}